\documentclass[10.5pt,reqno]{article}
\usepackage[margin=0.65in]{geometry} 
\usepackage[colorlinks,
            linkcolor=black,
            anchorcolor=black, 
            citecolor=black, 
            urlcolor=blue,
            ]{hyperref}     
\usepackage{indentfirst}
\usepackage{url}            
\usepackage{booktabs}       
\usepackage{amsfonts}       
\usepackage{nicefrac}       
\usepackage{microtype}      
\usepackage[reqno]{amsmath}
\usepackage{bbm}
\usepackage{graphicx}
\usepackage{subcaption}
\usepackage{wrapfig}
\usepackage{makecell}
\usepackage{enumitem}
\usepackage{multirow}
\usepackage[font=small,labelfont=bf]{caption}
\usepackage{colortbl}
\usepackage[scaled]{helvet}
\usepackage{courier} 
\usepackage[mathlines]{lineno}
\usepackage{bbm}
\usepackage{color}
\usepackage{xcolor}
\usepackage[most]{tcolorbox}
\usepackage{fvextra}
\usepackage{ulem}
\usepackage{amssymb}
\usepackage{nicematrix}

\newcommand{\kg}{\textsc{SciAtlas}}

\definecolor{aliceblue}{RGB}{178, 217, 245}

\definecolor{babyblue}{RGB}{217, 239, 251}

\usepackage[fixed]{fontawesome5}

\usepackage{textcomp}

\DefineVerbatimEnvironment{PromptVerbatim}{Verbatim}{
  breaklines=true,
  breakanywhere=true,
  breakindent=0pt,
  breaksymbolleft={},
  breaksymbolright={},
  fontsize=\small
}

\title{\vspace{-0.5cm}\large{\bf{SciAtlas: A Computable Atlas of Science for Knowledge-Grounded AI Research}}}

\author{
Shuofei Qiao \textsuperscript{1,2,*},
Yunxiang Wei \textsuperscript{1,*},
Busheng Zhang \textsuperscript{1,*},
Mengru Wang \textsuperscript{1,*},
Jiazheng Fan \textsuperscript{1,*},\\
Huadong Jian \textsuperscript{1},
Bin Wu \textsuperscript{2},
Shumin Deng \textsuperscript{1,$\dag$},
Yida Xue \textsuperscript{1},
Zifan Cheng \textsuperscript{1},
Xiang Chen \textsuperscript{1},\\
Dan Zhang \textsuperscript{4}, 
Junfeng Fang \textsuperscript{4},
Ningyu Zhang \textsuperscript{1,$\dag$},
Keyan Ding \textsuperscript{1},
Qiang Zhang \textsuperscript{1},\\
Jeff Z. Pan \textsuperscript{3},
Emine Yilmaz \textsuperscript{2},
Tat-Seng Chua \textsuperscript{4},
Huajun Chen \textsuperscript{1,$\dag$}
}

\date{}

\begin{document}
\maketitle

\vspace{-0.5cm}

{\renewcommand\baselinestretch{1.4}\selectfont
\centering
\noindent $^1$  Zhejiang University
$^2$ University College London\\
$^3$ University of Edinburgh
$^4$ National University of Singapore

\noindent $^*$ Equal contribution 
\noindent $\dag$ Corresponding Author

\texttt{\{shuofei,zhangningyu,huajunsir\}@zju.edu.cn}

\par}

\renewcommand{\figurename}{Fig.}

\flushbottom

\normalsize

\vspace{0.5cm}
{\centering\small
\faGithub~\textbf{Code:} \url{https://github.com/zjunlp/SciAtlas}\\
\faGlobe~\textbf{Project Website:} \url{http://sciatlas.openkg.cn}\par}
\vspace{-0.5cm}

\section*{Abstract}

{\renewcommand\baselinestretch{1.3}\selectfont

Artificial intelligence is rapidly entering the core workflows of scientific research. Yet reliable scientific reasoning requires access to accumulated scientific knowledge with sufficient breadth, depth, and standardization. Current AI scientists typically assemble scientific knowledge through workflow- and discipline-specific pipelines, which provide incomplete coverage, leave relations implicit, and make knowledge acquisition pathways fragmented. Here we present \textsc{SciAtlas}, a shared, machine-actionable cross-disciplinary scholarly knowledge infrastructure that integrates evidential, conceptual, disciplinary, expertise, and normative layers under a shared schema. This common schema makes heterogeneous discipline-specific scientific objects and relations accessible within the same machine-readable space. \textsc{SciAtlas} further achieves a unified neuro-symbolic retrieval mechanism that grounds heterogeneous research objects, propagates relevance across the scholarly topology, and projects the resulting relevance field into the context required by each scientific workflow, so that the same graph can recover different scientific objects and relations for different tasks through a common process. Across three representative workflows, \textsc{SciAtlas} broadens trajectory reconstruction by recovering overlooked stages and branches, deepens opportunity discovery by uncovering underexplored bottlenecks and cross-domain connections, and standardizes innovation assessment by integrating evidence, expertise, and evaluation signals. Case studies drawn from biology, chemistry and computer science illustrate how these capabilities operate in distinct research settings. Extensive evaluations on a curated benchmark demonstrate that \textsc{SciAtlas} achieves a nearly threefold improvement over the strongest baseline in latent-relation recovery and effectively improves multi-paper scientific synthesis on ScholarQABench, validating the foundational capabilities underpinning it as reusable knowledge infrastructure for knowledge-intensive scientific research.

\par}



\section{Main}

Artificial intelligence is rapidly entering the core workflows of scientific research.
Large language models and agentic systems can now coordinate scientific reasoning, planning, information seeking, and tool use across increasingly complex research processes, moving automated scientific research from concept toward practice \cite{ai4research-survey,ai-scientist-survey,ai-scientist,co-scientist,multi-agent-sci-dis}.

However, AI scientists cannot contribute reliably to research unless their reasoning, planning, and actions are grounded in the accumulated body of scientific knowledge \cite{ai-lack-creativity,auto-chem-llm}.
This knowledge supplies the evidence on which scientific claims rest, the relational structure through which inferences are made, and the evaluative standards against which scientific decisions are assessed.
This dependence places three requirements on how AI scientists access and use scientific knowledge:
1) \textit{Breadth}. AI scientists must situate a research problem within the wider development of its field, where relevant knowledge is dispersed across time and research communities.
Without such breadth, an agent may mistake the most readily accessible part of the literature for the structure of the field as a whole.
2) \textit{Depth}. AI scientists must move beyond topical relevance to uncover the underlying relations that connect scientific knowledge within and across disciplines.
Without such depth, an agent may reason on individually relevant studies yet remain unable to reconstruct the deeper scientific structure that gives those findings their collective meaning.
3) \textit{Standardization}. AI scientists must access and use scientific knowledge through a common machine-actionable framework that preserves the field-specific norms governing scientific practice.
Without such standardization, knowledge assembled for one discipline or workflow may not remain interpretable or reusable in another, fragmenting automated research into disconnected task-specific processes.

Despite rapid gains in agent capability, current AI scientists still struggle to access and use scientific knowledge with sufficient breadth, depth, and standardization.
Most obtain such knowledge through retrieval procedures designed around individual workflows \cite{openscholar,innoeval,surveyforge,chain-of-ideas,autosurvey,opennovelty}.
Even when expanded through agentic iteration, these procedures usually produce a transient collection of documents rather than a persistent representation of scientific knowledge \cite{ai-scientist,co-scientist,evoscientist,internagent}.
This design is poorly matched to science itself, whose meaning depends on relations distributed across the scholarly record rather than on documents considered in isolation.
The breadth available to the agent is bounded by the sources and formulations selected for the immediate task, leaving relevant regions of the scholarly landscape outside its view.
Depth is likewise limited because relations among retrieved findings remain implicit and must be reconstructed anew from isolated documents.
Standardization is difficult to achieve because what counts as valid evidence or accepted scientific practice differs across disciplines, while workflow-specific representations provide no common framework for encoding these differences in a form that agents can consistently access and apply.
These limitations share a common origin: scientific knowledge is treated as temporary task input rather than as a shared resource for automated research.
Addressing this bottleneck requires a machine-actionable infrastructure through which different agents and workflows can access the same organized body of scientific knowledge.

Knowledge graphs provide a natural basis for such an infrastructure because they represent scientific objects and their relations in a persistent, machine-actionable form.
Existing scientific knowledge graphs, however, are often bounded by particular disciplines, information objects, or schemas, limiting their use as a shared substrate for heterogeneous research workflows \cite{intern-atlas,matkg,biolink,openalex,open-research-kg,scholar-kg-review}.
Here we present \textsc{SciAtlas}, a large-scale scholarly knowledge graph designed as a common knowledge infrastructure for automated scientific research.
Spanning 26 disciplines, \textsc{SciAtlas} connects more than 157 million entities through 3 billion relational triples, including over 43 million papers, placing a broad scholarly record within the same machine-readable space.
It organizes this record through five interdependent layers under a unified schema.
The evidential layer establishes a paper-based backbone, upon which the conceptual layer represents fine-grained scientific concepts and their proximity across the literature.
Beyond this conceptual organization, the disciplinary layer situates papers within the hierarchy of science, while the expertise layer connects them to the communities that produce scientific knowledge.
The normative layer extends this structure to scientific evaluation by associating papers, where available, with review and evaluation signals, allowing differences in disciplinary standards to be represented within the shared schema.
Together, these layers make the deeper organization of scientific knowledge explicit, tracing how it develops and interacts within and across fields.
A unified neuro-symbolic retrieval mechanism then grounds heterogeneous research objects as seed nodes and propagates relevance across graph relations, allowing different agents and workflows to recover the knowledge they require from the same scholarly substrate.

We evaluate \textsc{SciAtlas} across three representative scientific workflows.
In research development trajectory, \textsc{SciAtlas} enables agents to recover broader field histories and more complete methodological structures.
In research opportunity discovery, it supports deeper formulation by locating unresolved problems in the development of a field and connecting them to mechanisms beyond the immediate domain.
In innovation assessment, \textsc{SciAtlas} makes field-specific evidence and evaluative norms accessible through a common representation, supporting more grounded judgement and more actionable revision.
To determine whether these gains arise from a shared capability rather than workflow-specific design, we construct \textsc{SciAtlasEval}, a self-curated evaluation suite for latent-relation recovery and task-relevant scientific-object retrieval from heterogeneous research inputs.
\textsc{SciAtlas} achieves nearly a threefold improvement in nDCG@20 over the strongest baseline for latent-relation recovery, while more effectively identifying the papers and experts needed to contextualize diverse research objects.
On an external evaluation of multi-paper scientific synthesis, its graph-structured context helps integrate evidence distributed across connected studies into more comprehensive and better-grounded scientific answers.
Together, these results establish \textsc{SciAtlas} as a shared, machine-actionable knowledge infrastructure through which diverse research workflows can access and reason over scientific knowledge using a common mechanism.

\begin{figure*}[!th] 
  \vspace{-0.7cm}
  \centering
  \includegraphics[width=1\linewidth]{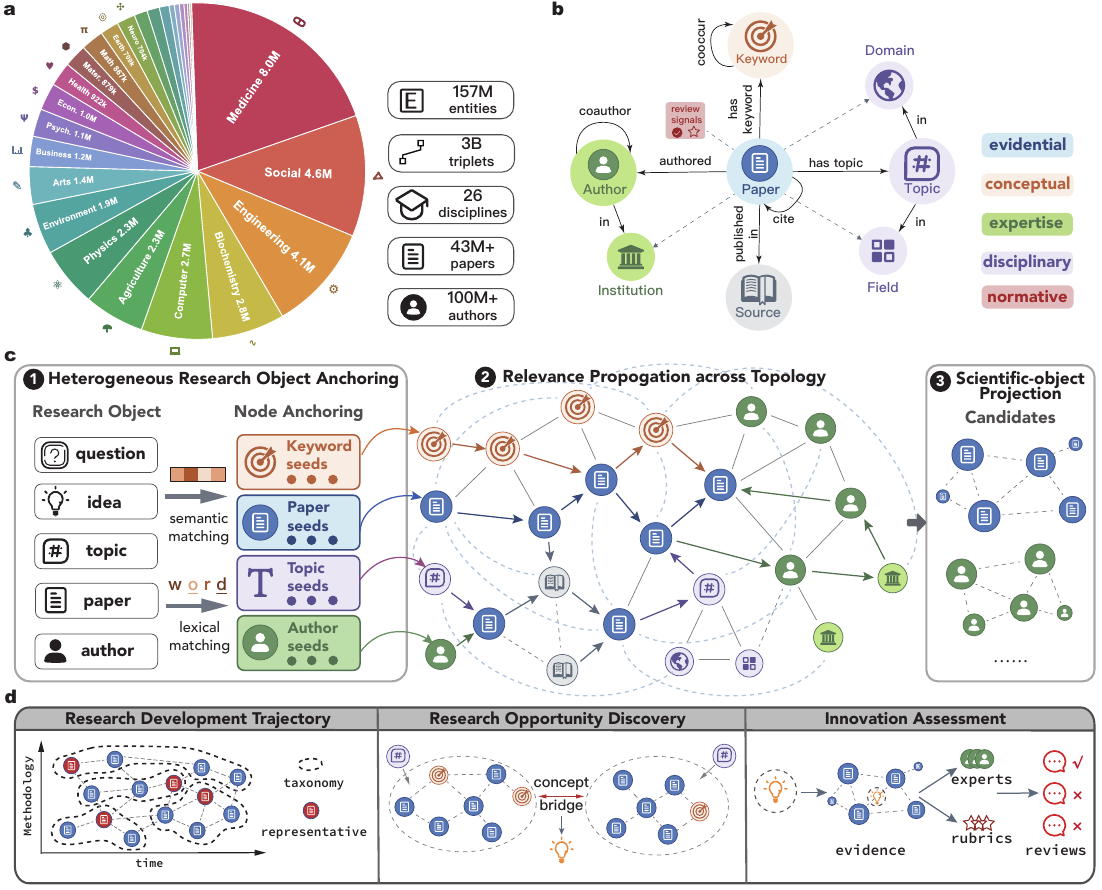}
  \caption{
  \small
  \textbf{Overview of {\kg}.}
  \textbf{a,} Scale of {\kg}. {\kg} contains more than 157M entities and 3B triples, including 43M papers and 100M authors, and spans 26 disciplines across science, engineering, etc.
  \textbf{b,} Scholarly topology of {\kg}. Centred on papers, {\kg} organizes heterogeneous scholarly knowledge into five interdependent layers encompassing evidence, concepts, disciplines, expertise, and evaluation signals.
  \textbf{c,} Unified neural-symbolic retrieval. Heterogeneous research objects are grounded in the scholarly graph through seed nodes, from which relevance is propagated across the topology to recover latent scientific context beyond document similarity. \textsc{SciAtlas} and this unified retrieval mechanism are deployed as accessible services for practical use.
  \textbf{d,} Workflow-specific context assembly. The shared relevance field is projected into task-specific structures for research development trajectory reconstruction, research opportunity discovery, and innovation assessment. These three workflows are also deployed as services and agentic skills that can be used across diverse agent frameworks (e.g., Codex, Claude Code, OpenClaw). 
  }
  \vspace{-0.2cm}
  \label{fig:overview}
\end{figure*}

\section{Results}

\begin{figure*}[!th] 
  \vspace{-0.7cm}
  \centering
  \includegraphics[width=.95\linewidth]{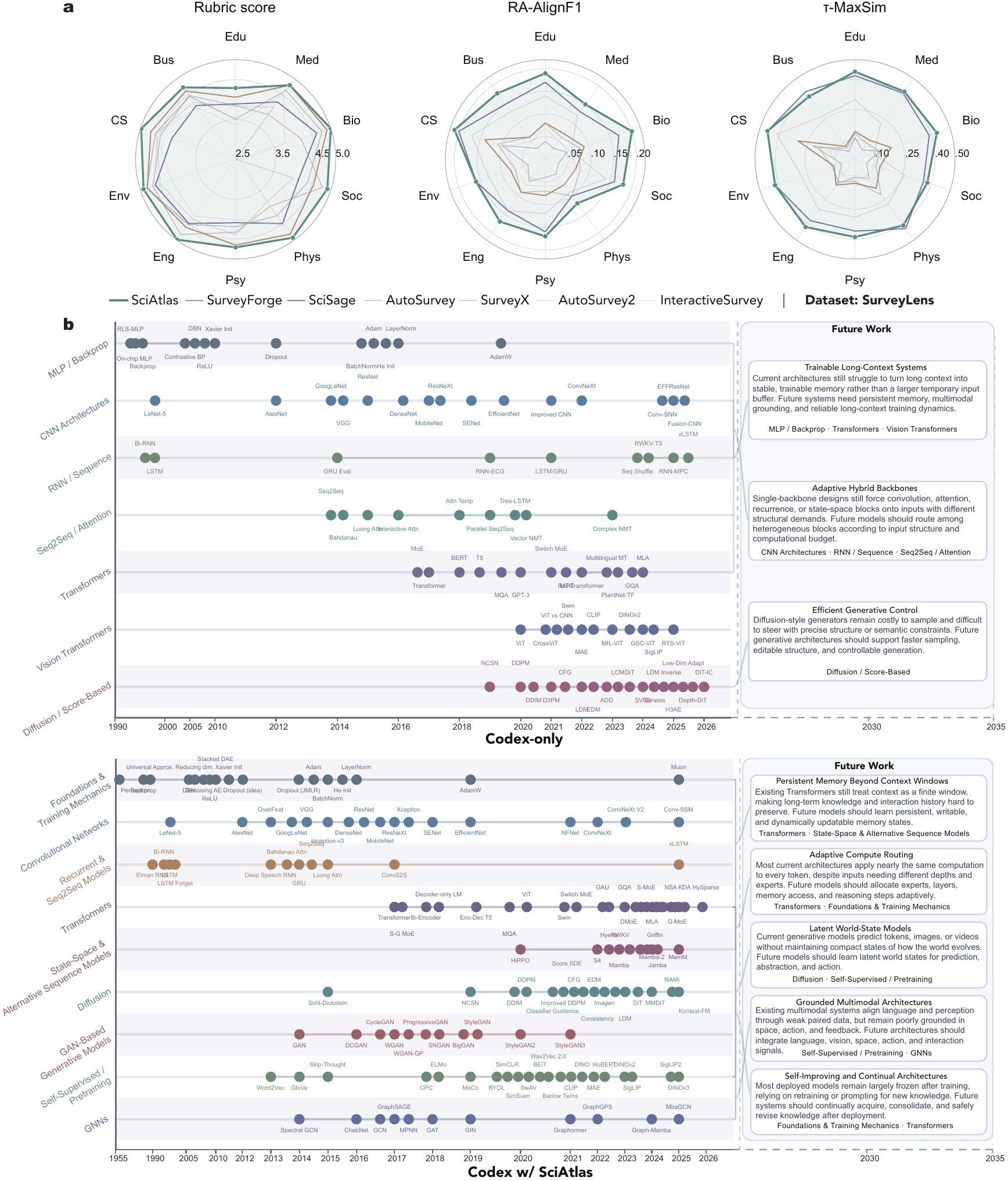}
  \caption{
    \small
    \textbf{Quantitative evaluation and controlled breadth analysis of Research Development Trajectory reconstruction.}
    \textbf{a,} Performance on SurveyLens \cite{surveylens} across ten disciplines.
    The radar plots report Rubric Score, RA-AlignF1, and $\tau$-MaxSim for {\kg}-assisted RDT and six agentic literature-review synthesis baselines.
    Each spoke denotes one discipline, and larger radial values indicate better performance.
    \textbf{b,} Controlled comparison on deep neural network architecture development.
    Codex-only (top) and Codex with {\kg} (bottom) use the same \textsc{GPT-5.5} backbone and RDT workflow, differing only in the retrieval substrate.
    Points denote representative studies positioned by publication year, horizontal rows group them into methodological branches, and the boxes on the right show future-work directions synthesized from each reconstructed trajectory.
    }
  \vspace{-0.5cm}
  \label{fig:literature_review-1}
\end{figure*}

\subsection{Overview of {\kg}}

We develop {\kg} as a shared, machine-actionable, cross-disciplinary scholarly knowledge infrastructure for automated scientific research.
It organizes heterogeneous scientific objects and their relations within a common graph that spans 26 disciplines, comprising more than 43 million papers, 157 million entities, and 3 billion relational triples (Fig.~\ref{fig:overview}a).
Centred on scientific papers, {\kg} integrates five complementary layers under a shared schema (Fig.~\ref{fig:overview}b).
The evidential layer provides the paper-centred backbone through citation and publication relations.
We specifically construct the conceptual layer containing normalized keyword nodes extracted from papers, paper--keyword associations, and keyword co-occurrence edges to bridge implicit relations among scientific objects, connect equivalent or related concepts expressed through different terminology and levels of abstraction across disciplines, and provide fine-grained anchors for retrieval beyond direct document similarity.
The disciplinary layer situates papers within hierarchical research fields, while the expertise layer connects scholarly records to the researchers, institutions, and communities that produce them.
We also include a normative layer to further associate papers with their review signals, if available, to reflect field-specific evaluation norms.
Together, these layers place scientific evidence, concepts, disciplines, expertise, and evaluation signals within a single machine-readable scholarly topology.
To access this topology, {\kg} uses a unified neuro-symbolic retrieval to ground heterogeneous research objects, propagate relevance across scholarly relations, and project the resulting relevance field into workflow-specific context (Fig.~\ref{fig:overview}c).
Given a research object, symbolic and neural anchoring map explicitly named and underspecified inputs, respectively, onto heterogeneous seed nodes.
Relevance then propagates along relation-aware paths, allowing support from connected nodes to accumulate beyond direct matches.
The resulting field is projected into the scientific objects and relations required by each workflow, allowing task-specific outputs to be obtained through the same graph and retrieval process.
\textsc{SciAtlas} and its unified retrieval mechanism are deployed as accessible services, enabling agents to access this shared scholarly topology in practical research settings.

Across three representative workflows, we examine how this shared capability supports distinct scientific objectives (Fig.~\ref{fig:overview}d).
For research development trajectory reconstruction, {\kg} converts the distributed scholarly record into a structured account of how a field evolves.
For research opportunity discovery, it reveals underexplored positions in the evolving structure of a field and connects them to mechanisms that may open new research directions.
For innovation assessment, it situates a proposed idea within the scholarly context that shapes how the work should be judged and improved.
Coupled with LLM agents, these contextual views support distinct forms of scientific reasoning while remaining grounded in the same scholarly substrate.
The three workflows are also deployed as reusable services and agentic skills, allowing diverse agent frameworks to invoke the same context-assembly process.
Thus, {\kg} provides not a separate knowledge pipeline for each task, but a common infrastructure from which different scientific workflows and agentic frameworks can assemble the context they require.

\subsection{{\kg} broadens research development trajectory reconstruction}
Research Development Trajectory (RDT) requires an agent to recover more than relevant papers.
It must identify the developmental stages, methodological branches, representative evidence, and conceptual transitions that explain how a scientific field evolves.
This makes RDT a direct test of breadth, because a useful trajectory must extend beyond prominent milestones to less visible periods and branches.
To support RDT, {\kg} projects retrieved papers and relations into temporal and methodological structures, with taxonomies and representative studies marking developmental branches.
To quantitatively assess {\kg}-assisted RDT across topics and disciplines, we use survey generation as a proxy evaluation task because scientific surveys likewise synthesize how a field develops, organizing its stages, methodological branches, and representative evidence into a coherent account.

\paragraph{Experimental setup}
We use SurveyLens \cite{surveylens}, a discipline-aware dataset for automatic survey generation, which contains 1,000 human-written surveys across 10 disciplines.
We compare {\kg} with six agentic literature-review synthesis systems: AutoSurvey \cite{autosurvey}, SurveyForge \cite{surveyforge}, AutoSurvey2 \cite{autosurvey2}, InteractiveSurvey \cite{interactivesurvey}, SurveyX \cite{surveyx}, and SciSage \cite{scisage}.
All agentic components use \textsc{GPT-5.5} as the LLM backbone.
For evaluation metrics, the Rubric Score assesses outline organization, content synthesis, and reference quality; RA-AlignF1 measures redundancy-aware coverage and faithfulness, whereas $\tau$-MaxSim measures semantic alignment with the reference survey.
Full metric definitions and baseline settings are provided in Appx.~\ref{app:metrics-surveylens} and \ref{app:baselines-rdt}.

\begin{figure*}[!th] 
  \vspace{-0.5cm}
  \centering
  \includegraphics[width=1.\linewidth]{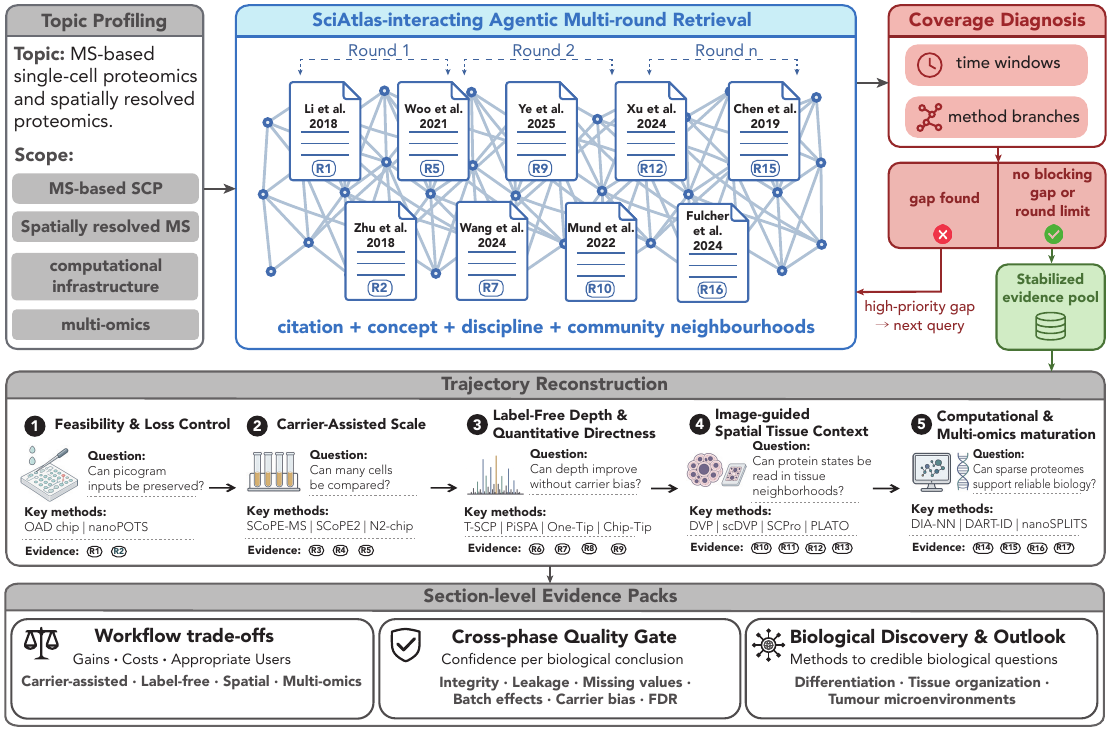}
  \caption{
\small
\textbf{End-to-end Research Development Trajectory reconstruction for mass-spectrometry-based single-cell and spatially resolved proteomics.}
Topic profiling defines the field boundary across MS-based single-cell proteomics, spatially resolved mass spectrometry, computational infrastructure, and multi-omics.
The {\kg}-interacting agent retrieves evidence over multiple rounds across citation, concept, discipline and community neighbourhoods.
Coverage diagnosis examines missing time windows and methodological branches; high-priority gaps trigger targeted follow-up queries until no blocking gap remains or the round limit is reached.
The stabilized evidence pool is organized into five overlapping developmental shifts: Feasibility and Loss Control, Carrier-Assisted Scale, Label-Free Depth and Quantitative Directness, Image-guided Spatial Tissue Context, and Computational and Multi-omics Maturation.
Each shift presents its guiding question, key methods, and evidence identifiers, linking the reconstructed trajectory to representative studies.
The five shifts are further projected into three section-level evidence packs: Workflow Trade-offs compares gains, costs, and appropriate uses; the Cross-phase Quality Gate consolidates criteria for confidence per biological conclusion; and Biological Discovery and Outlook connects methodological advances to credible biological questions.
}
  \vspace{-0.5cm}
  \label{fig:literature_review-2}
\end{figure*}

\paragraph{Results}
{\kg} achieves the strongest overall performance across all three metrics (Fig.~\ref{fig:literature_review-1}a).
Its macro-average Rubric Score reached 4.82, compared with 4.67 for the strongest baseline, SurveyForge.
It matches or exceeds the best baseline in all ten disciplines, showing that the improvement was consistent across fields.
{\kg} also achieves the highest RA-AlignF1 and mean $\tau$-MaxSim scores, indicating stronger non-redundant coverage of human survey entries and closer semantic alignment with their content.
Together, these results show that {\kg}-assisted RDT produces surveys that more closely follow discipline-specific review standards and more faithfully recover the content of human-written reviews.
Stronger conformity to discipline-specific review standards reflects how {\kg} structures retrieved evidence into field-specific research threads, developmental stages, and evidence roles before synthesis.
Closer agreement with human-written surveys reflects the unified schema of {\kg} and its topological retrieval, which expands the evidence pool and enables coverage diagnosis to recover foundational periods, methodological branches, and representative studies beyond locally salient papers.
To further validate the breadth of {\kg}-assisted trajectory reconstruction, we conduct two case studies: a controlled comparison of deep neural network architecture development and an end-to-end reconstruction of mass-spectrometry-based single-cell and spatially resolved proteomics.

\paragraph{Controlled comparison of trajectory breadth}
To isolate how the retrieval substrate affects RDT breadth, we use DNN architecture development, a field with a long history and dense methodological branching.
We compare Codex with and without {\kg}; both settings use \textsc{GPT-5.5} and the same RDT workflow, differing only in evidence retrieval.
Both settings recover the principal architecture lineages (Fig.~\ref{fig:literature_review-1}b).
However, the Codex-only reconstruction concentrates on highly visible milestones, whereas {\kg} additionally recovers GAN-based models, self-supervised pretraining, graph neural networks, and state-space models.
It also identifies more representative works within the lineages recovered by both settings.
These additions transform the output from a milestone-centred timeline into a broader field map that exposes how methodological traditions branch and converge.
The wider historical structure also changes future-work synthesis, linking current bottlenecks to multiple branches and unresolved transitions rather than only dominant recent lineages.
This difference arises because topological retrieval reaches predecessor works, adjacent methodological families, and lexically distant branches, giving coverage diagnosis a relation-grounded view of missing periods and underrepresented threads.
Under the same reasoning workflow, this controlled case shows that {\kg} broadens trajectory reconstruction by recovering field-level structure beyond locally salient papers.

\begin{figure*}[!th] 
  \vspace{-0.5cm}
  \centering
  \includegraphics[width=1.\linewidth]{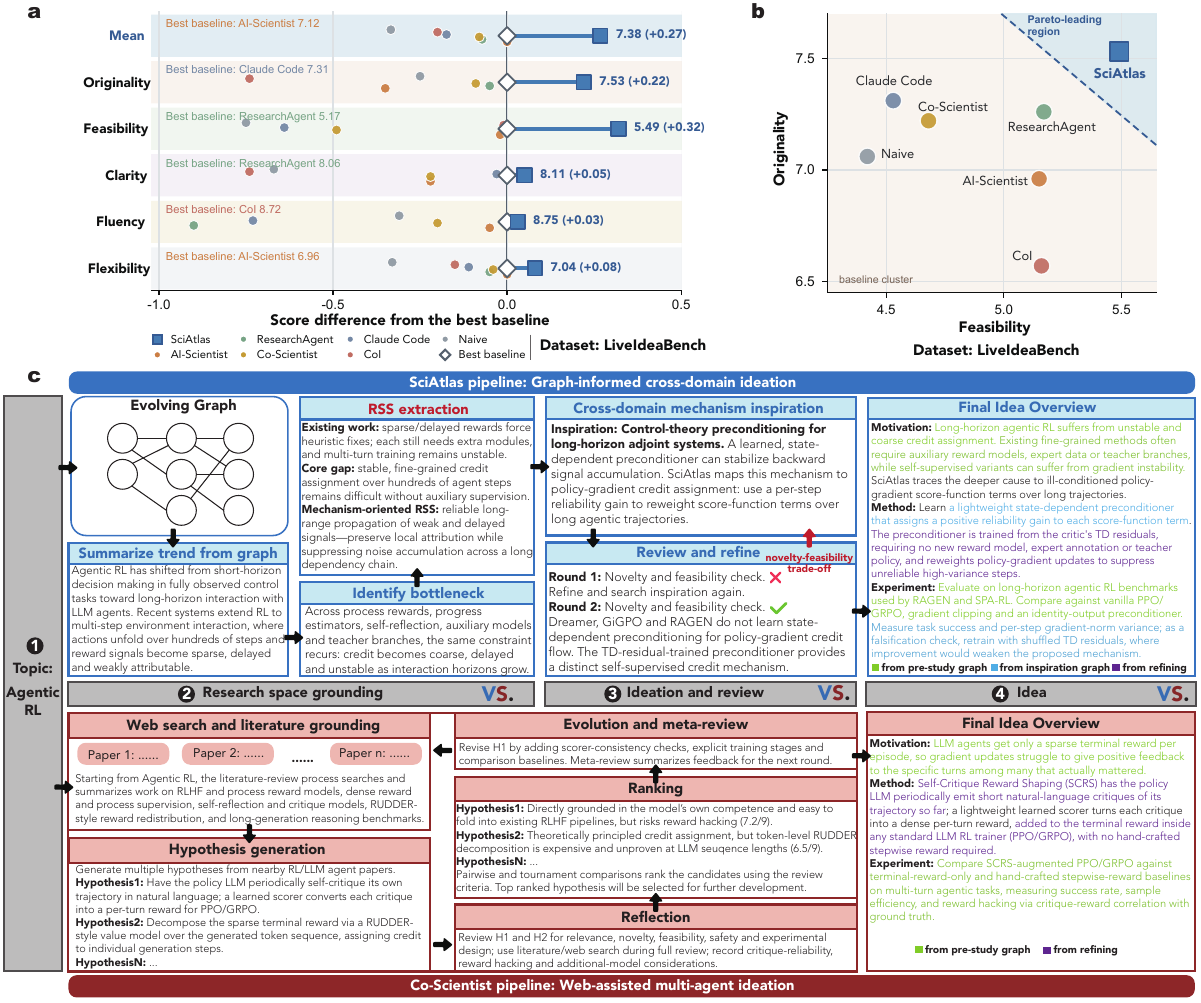}
  \caption{
    \small
    \textbf{Quantitative performance and comparative depth analysis of Research Opportunity Discovery.}
    \textbf{a,} Performance on LiveIdeaBench \cite{liveideabench} across the overall mean, originality, feasibility, clarity, fluency, and flexibility.
    Scores are centred on the strongest baseline in each row (diamond, zero); circles denote the remaining baselines and blue squares denote {\kg}.
    Labels report the absolute {\kg} score and its improvement over the strongest baseline.
    \textbf{b,} Originality--feasibility landscape of the evaluated systems.
    The shaded area above the dashed boundary denotes the Pareto-leading region, where {\kg} combines the highest originality and feasibility.
    \textbf{c,} Comparative depth analysis in agentic reinforcement learning under the same topic input and \textsc{GPT-5.5} backbone.
    {\kg} reconstructs the shift towards long-horizon interaction, identifies unstable fine-grained credit assignment as the field-level bottleneck and maps control-theory preconditioning into a state-dependent policy-gradient mechanism.
    Co-Scientist instead uses literature grounding and iterative multi-agent refinement to develop self-critique reward shaping from neighbouring RL and LLM-agent research.
    The {\kg} proposal further connects its transferred mechanism to concrete baselines and mechanism-matched ablations, whereas Co-Scientist provides broader task-level comparisons.
    }
  \vspace{-0.5cm}
  \label{fig:oppo_dis-1}
\end{figure*}

\paragraph{End-to-end trajectory reconstruction in single-cell and spatial proteomics}
We use mass-spectrometry-based single-cell and spatially resolved proteomics to test whether the breadth recovered by {\kg} can be converted into a complete and scientifically useful field trajectory (Fig.~\ref{fig:literature_review-2}).
This field develops through coupled trade-offs among sensitivity, proteome depth, throughput, quantitative accuracy, spatial context, and biological interpretability rather than along a single performance axis.
The reconstructed trajectory resolves five overlapping developmental shifts: loss control \cite{oil-air-droplet,nanodroplet}; carrier-assisted scale \cite{scope-ms,scope2}; label-free depth and accuracy \cite{t-scp,pispa,one-tip,chip-tip}; spatial tissue context \cite{dvp,scdvp,scpro,plato}; and computational, quality-control, and multi-omic maturation \cite{dia-nn,dart-id,nanosplits}.
Rather than treating these shifts as a linear chronology, {\kg} organizes representative evidence into complementary views of workflow trade-offs, cross-phase quality control, and biological discovery.
The trade-off view compares the gains, costs, and appropriate uses of carrier-assisted, label-free, spatial, and multi-omic designs.
The quality gate shifts evaluation from maximizing proteins per cell to increasing confidence per biological conclusion \cite{rec-bench-single-cell,benchmarking-informatics}.
The biological-discovery view connects these technical advances to protein-level studies of differentiation, tissue organization and tumour microenvironments \cite{blood-cell,scdvp,scpro}.
Together, this case shows that the breadth enabled by {\kg} is not simply greater paper coverage.
By recovering developmental stages and methodological branches, {\kg} connects methods and bottlenecks to evidence standards and credible biological questions in a paper-grounded, auditable trajectory.

\subsection{{\kg} deepens research opportunity discovery}

\begin{figure*}[!th] 
  \vspace{-0.5cm}
  \centering
  \includegraphics[width=1.\linewidth]{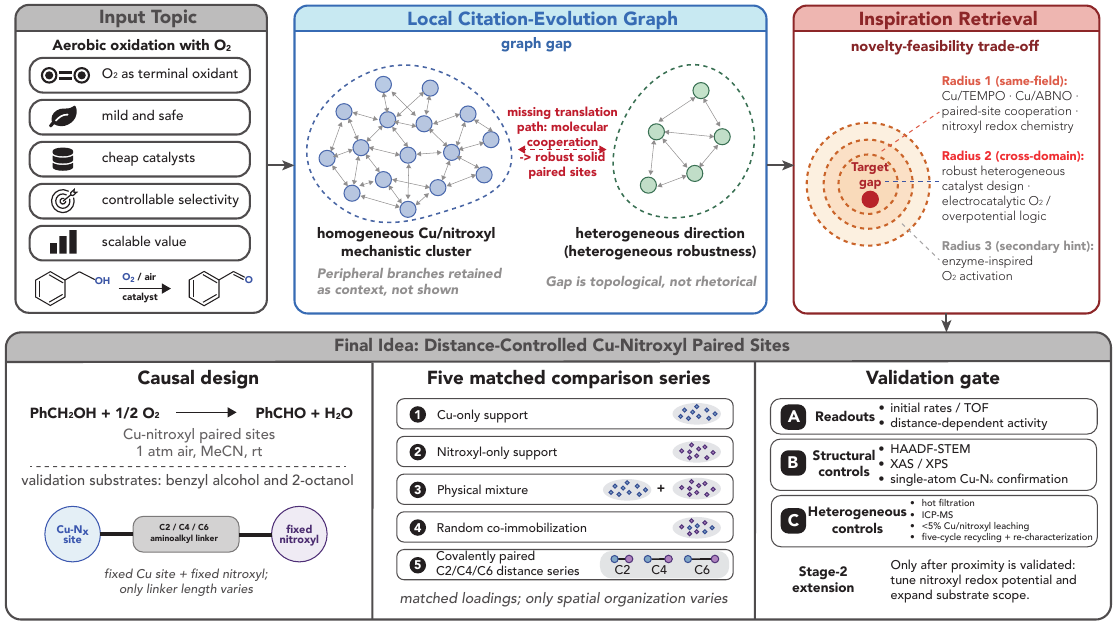}
  \caption{
\small
\textbf{End-to-end opportunity discovery in aerobic oxidation.}
Starting from aerobic oxidation with O$_2$ or air, the local citation-evolution graph reveals a missing translation path between homogeneous Cu/nitroxyl mechanisms and robust heterogeneous paired sites.
Inspiration retrieval combines same-field Cu/nitroxyl chemistry with cross-domain principles for heterogeneous-catalyst robustness and redox control.
Mechanism mapping produces a distance-controlled Cu--nitroxyl catalyst that fixes the Cu site, nitroxyl identity, support, and anchoring chemistry while varying only the C2/C4/C6 linker length.
Five matched comparison series and validation gates test paired-site proximity, structural assignment, and heterogeneous stability before further redox tuning and substrate expansion.
The figure shows how {\kg} converts a topological gap into a one-variable, actionable, and falsifiable experimental proposal.
}
  \vspace{-0.3cm}
  \label{fig:oppo_dis-2}
\end{figure*}

Finding a research opportunity involves more than extrapolating from the nearest relevant literature.
Research Opportunity Discovery (ROD) locates unresolved bottlenecks in an evolving field, searches for mechanisms that may address them, and translates the resulting connections into testable directions.
Because gaps and transferable mechanisms are defined by relations among research threads, ROD depends on more than local evidence.
Its depth lies in progressing from field-level diagnosis to cross-domain connection and explicit validation.
Within the {\kg}-assisted workflow, retrieved papers and relations form a local citation-evolution graph.
A Research Structure Signature distils its unresolved bottlenecks and guides the retrieval of structurally matched mechanisms from same-field and cross-domain neighbourhoods.
For large-scale evaluation across topics and disciplines, scientific idea generation serves as a proxy because the resulting ideas reveal whether these gaps and mechanisms support original yet feasible research directions.

\paragraph{Experimental setup}
We use LiveIdeaBench \cite{liveideabench}, which evaluates scientific idea generation from 1,180 high-impact keywords spanning 22 scientific domains.
We compare {\kg} with six scientific idea-generation baselines: Naive LLM, Chain-of-Ideas \cite{chain-of-ideas}, ResearchAgent \cite{researchagent}, AI-Scientist \cite{ai-scientist}, Co-Scientist \cite{co-scientist}, and Claude Code.
All agentic components use \textsc{GPT-5.5} as the LLM backbone.
Following LiveIdeaBench, originality, feasibility, and clarity are averaged over three judges randomly sampled from a diverse ten-model panel for each idea.
Fluency is evaluated by one randomly sampled judge for each keyword--system pair based on the diversity of its generated idea set.
Flexibility is computed as the 30th percentile of keyword-level composite scores.
We average their scores within each dimension and report both overall and dimension-level performance.
Full metric definitions and baseline settings are provided in Appx.~\ref{app:metrics-liveideabench} and \ref{app:baselines-opportunity}.

\paragraph{Results}
{\kg}-assisted ROD achieves the strongest overall performance on LiveIdeaBench, with a mean score of 7.38 compared with 7.12 for the strongest baseline, AI-Scientist (Fig.~\ref{fig:oppo_dis-1}a).
It ranks first in all five dimensions.
Its simultaneous gains in originality and feasibility place {\kg} in the Pareto-leading region of the originality--feasibility landscape (Fig.~\ref{fig:oppo_dis-1}b).
The originality advantage reflects how the local citation-evolution graph and Research Structure Signature expose field-level bottlenecks, while topological retrieval introduces structurally matched mechanisms beyond the immediate domain.
Feasibility improves as mechanism mapping and creativity--feasibility feedback translate these inspirations into target-specific mechanisms and explicit validation paths.
The gains in clarity and fluency reflect complementary aspects of the same structure: the Research Structure Signature organizes each direction into a coherent proposal, whereas multi-radius retrieval supplies diverse, non-redundant mechanism families.
The leading flexibility further indicates that this knowledge-acquisition process transfers across topics and disciplines.
Although the target evidence changes, the shared schema and topological retrieval provide a consistent route for acquiring and organizing it.
Together, these results show that {\kg} improves opportunity formulation at scale by connecting underexplored field positions to original, feasible, and consistently generated research directions.
To examine where this depth arises and how it carries into an experimentally specified proposal, we compare {\kg} with Co-Scientist in agentic reinforcement learning and present an end-to-end case in aerobic oxidation.

\paragraph{Comparative depth analysis in agentic reinforcement learning}
We next examine agentic reinforcement learning (RL), a rapidly developing area at the intersection of reinforcement learning and LLM agents (Fig.~\ref{fig:oppo_dis-1}c).
We compare {\kg} with Co-Scientist under the same topic input and \textsc{GPT-5.5} backbone.
{\kg} reconstructs a field-wide shift from short-horizon decision-making to long-horizon LLM-agent interaction and consolidates recurring limitations into unstable, fine-grained credit assignment.
Co-Scientist instead frames the opportunity around sparse terminal rewards, leaving the broader field-level bottleneck implicit.
The resulting Research Structure Signature abstracts this bottleneck as a long-horizon signal-propagation problem, allowing topological retrieval to identify control-theory preconditioning as a structurally matched cross-domain mechanism.
Mechanism mapping and creativity--feasibility feedback then convert this principle into a state-dependent preconditioner that reweights policy-gradient credit using critic TD residuals.
By contrast, Co-Scientist develops self-critique reward shaping from mechanisms within the neighbouring RL and LLM-agent literature.
At the validation level, Co-Scientist proposes broad comparisons against terminal-reward and hand-crafted reward baselines, whereas {\kg} specifies concrete baseline methods and mechanism-matched ablations to isolate whether the observed gains arise from preconditioning itself.
Because {\kg} preserves the link between the diagnosed credit-assignment bottleneck and the transferred preconditioning mechanism, it can translate each scientific claim into a corresponding baseline or ablation, yielding a more actionable and falsifiable experimental plan.

\paragraph{End-to-end opportunity discovery in aerobic oxidation}
We use aerobic oxidation to examine whether the depth enabled by {\kg} can be converted into a concrete and experimentally testable research opportunity (Fig.~\ref{fig:oppo_dis-2}).
The local citation-evolution graph reveals a dense homogeneous Cu/nitroxyl mechanistic cluster but a weak connection to heterogeneous catalysis, exposing a missing translation path from molecular cooperation to stable solid paired sites.
The Research Structure Signature formulates this discontinuity as a specific bottleneck: transferring Cu/nitroxyl cooperation into recyclable solids while disentangling spatial proximity, redox coupling, and leaching resistance.
Guided by this structure, same-field retrieval preserves mechanistic continuity with Cu/TEMPO, Cu/ABNO and nitroxyl redox chemistry \cite{h-itempo,m-itempo,copper-iabno,org-oxi,ele-alc-oxi}, whereas cross-domain retrieval introduces principles for robust heterogeneous-site design and redox control \cite{robust-heter,ele-alc-oxi}.
Mechanism mapping converts these connections into a distance-controlled Cu--nitroxyl paired-site catalyst comprising single-atom Cu sites in N-rich g-C$_3$N$_4$ pockets and covalently tethered nitroxyl co-catalysts.
The proposal fixes the Cu site, nitroxyl identity, support, and anchoring chemistry while varying only the Cu--nitroxyl distance through C2/C4/C6 linkers.
Cu-only, nitroxyl-only, physical-mixture, and randomly co-immobilized controls test whether activity arises specifically from paired-site proximity, while hot-filtration, ICP-MS, and recycling tests examine heterogeneous stability.
By preserving the connection from graph-grounded bottleneck to transferred mechanism and matched controls, {\kg} turns a broad topic into a traceable, actionable and falsifiable research opportunity.

\begin{figure*}[!th] 
  \vspace{-0.5cm}
  \centering
  \includegraphics[width=.97\linewidth]{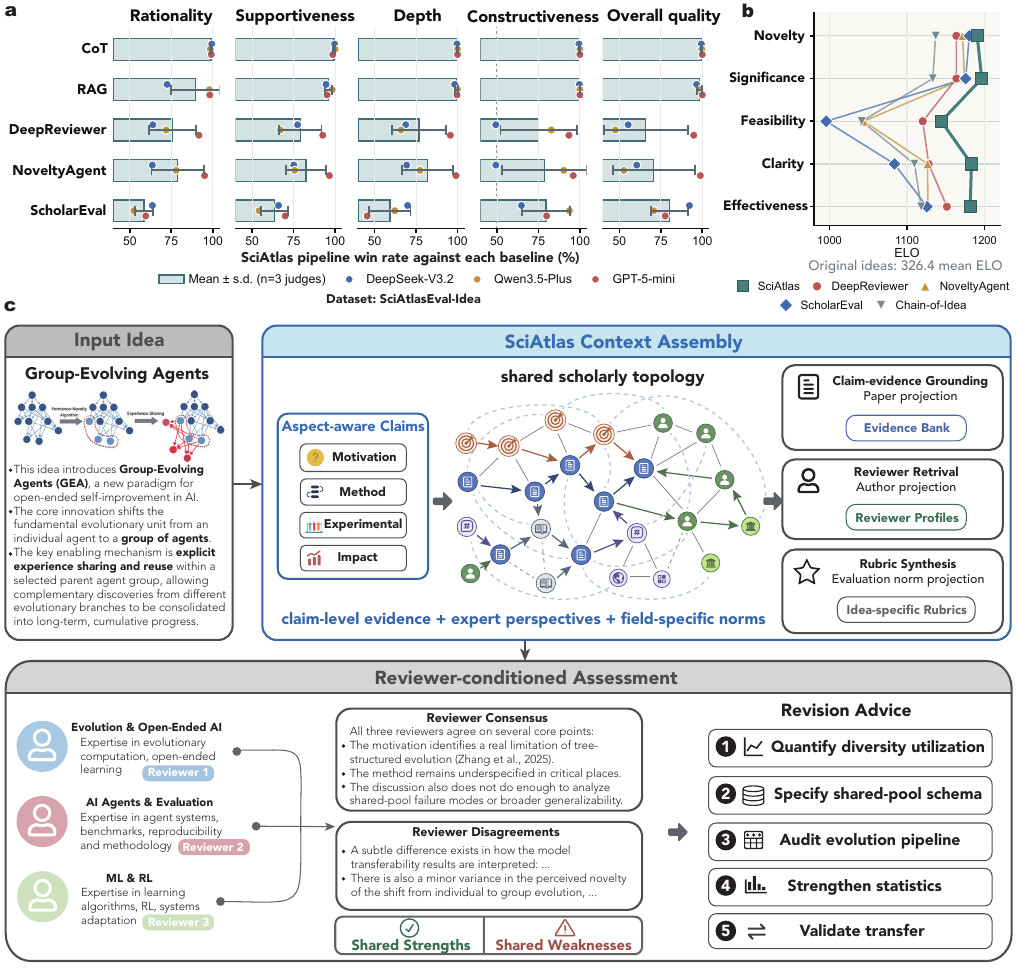}
  \caption{
\small
\textbf{Quantitative evaluation and end-to-end analysis of {\kg}-assisted Innovation Assessment.}
\textbf{a,} Pairwise assessment-report evaluation on \textsc{SciAtlasEval-Idea} across rationality, supportiveness, depth, constructiveness, and overall quality.
Bars show the mean win rate of {\kg} against CoT, RAG, DeepReviewer, NoveltyAgent and ScholarEval, with standard deviations and judge-specific results from three LLM judges.
\textbf{b,} Feedback-guided idea refinement.
Ideas revised using feedback from {\kg} and the assessment baselines are ranked by ELO across novelty, significance, feasibility, clarity, and effectiveness; the annotation reports the mean ELO of the unoptimized original ideas.
\textbf{c,} End-to-end assessment of \textit{Group-Evolving Agents} \cite{group-evolving}.
{\kg} decomposes the input into aspect-aware claims and projects the shared scholarly topology into a claim-level evidence bank, publication-grounded reviewer profiles, and idea-specific rubrics.
These contexts condition reviewer-specific assessments whose consensus and disagreements are consolidated into shared strengths, weaknesses, and revision advice.
The displayed findings and revision targets are representative excerpts from the complete assessment.
The panel shows how a common context-assembly process connects idea-specific evidence, expert perspectives, and field-specific norms to traceable judgment and actionable revision.
}
  \vspace{-0.5cm}
  \label{fig:inno_ass}
\end{figure*}

\subsection{{\kg} standardizes innovation assessment}

Innovation Assessment (IA) goes beyond assigning novelty or plausibility scores: it requires establishing novelty boundaries, examining claim support, and identifying revisions using relevant evidence, expert perspectives, and field-specific evaluation norms.
Because these resources vary across ideas and disciplines, current systems typically assemble them through separate task-specific procedures.
IA therefore tests standardization not through uniform criteria, but through a common process for acquiring and organizing field-specific assessment context.
Within the {\kg}-assisted workflow, assessable claims guide the construction of a claim-level evidence bank, publication-grounded reviewer profiles, and idea-specific rubrics, which jointly support reviewer-specific assessments, meta-review, and revision advice.
We evaluate this capability through assessment-report quality and the downstream improvement produced by its feedback.

\paragraph{Experimental setup}
We evaluate {\kg}-assisted IA at two levels: assessment-report quality and feedback-guided idea refinement.
For report quality, we use \textsc{SciAtlasEval-Idea}, the idea-level component of our curated benchmark suite for heterogeneous scientific-object retrieval and relation recovery, comprising 349 paper-grounded research ideas spanning published, rejected, and unpublished work.
We compare {\kg} with five assessment baselines: chain-of-thought prompting (CoT), retrieval-augmented generation (RAG), DeepReviewer \cite{deepreview}, NoveltyAgent \cite{noveltyagent} and ScholarEval \cite{scholareval}.
Each {\kg} report is paired with the corresponding baseline report and independently evaluated by DeepSeek-V3.2, Qwen3.5-Plus, and GPT-5-mini across rationality, supportiveness, depth, constructiveness, and overall quality.
We report the mean win rate of {\kg} across ideas and judges.
For downstream refinement, we use the unoptimized Chain-of-Ideas outputs from 50 recent AI research topics as the original ideas \cite{chain-of-ideas}.
Each idea is revised using feedback from {\kg}, DeepReviewer, NoveltyAgent, or ScholarEval, with the standard Chain-of-Ideas refinement included as an additional baseline.
The refined and original ideas are evaluated through reversed-order pairwise comparisons and converted into ELO scores for novelty, significance, feasibility, clarity, and effectiveness.
All configurable agentic components use \textsc{GPT-5.5}.
Full metric definitions and baseline settings are provided in Appx.~\ref{app:metrics-idea-ass}, \ref{app:metrics-idea-refinement} and \ref{app:baselines-inno-ass}.

\paragraph{Results}
{\kg}-assisted IA produces assessment reports that are consistently preferred over all evaluated baselines (Fig.~\ref{fig:inno_ass}a).
Its mean win rates reach 99.6\% against CoT and 96.6\% against RAG, and remain substantial against the specialized systems DeepReviewer (74.5\%), NoveltyAgent (78.5\%) and ScholarEval (68.2\%).
The preference spans rationality, supportiveness, depth, constructiveness, and overall quality, indicating that the advantage is not confined to a single aspect of assessment.
This pattern reflects how {\kg} constructs assessment context.
Whereas the baselines operate on the idea text, a flat evidence set, or a specialized evaluation perspective, {\kg} organizes claim-level evidence, publication-grounded reviewer profiles, and idea-specific rubrics around the same assessable claims.
Although the retrieved evidence, experts, and evaluation criteria vary across ideas, the shared schema provides a consistent process for assembling and applying them.
Each judgment can therefore be traced to a claim, comparison set, expert perspective or expected form of evidence.
This advantage carries into downstream idea refinement (Fig.~\ref{fig:inno_ass}b).
Ideas revised using {\kg}-generated feedback achieve the highest ELO across all five criteria, with a mean of 1179.2 compared with 1145.5 for the strongest baseline, DeepReviewer, and 326.4 for the unoptimized original ideas.
The largest margins over the strongest baseline occur in clarity (+55.2), effectiveness (+30.1) and feasibility (+23.9).
Claim-level grounding identifies unsupported or overextended assertions, while reviewer perspectives and idea-specific rubrics translate these weaknesses into targets for additional evidence, methodological specification, and experimental validation.
The same structured context therefore connects scientific judgment directly to revision, explaining why the feedback improves not only how ideas are presented but also how concretely they can be evaluated and implemented.
Together, the report-quality and refinement results show that {\kg} standardizes how idea-specific evidence, expertise, and evaluation norms are converted into traceable judgments and actionable revisions.
To examine how this process operates within a complete assessment, we apply it to a detailed case study.

\paragraph{End-to-end assessment of Group-Evolving Agents}
We use the idea of \textit{Group-Evolving Agents} \cite{group-evolving} to trace how {\kg} carries standardized context assembly through a complete IA workflow (Fig.~\ref{fig:inno_ass}c).
The idea shifts open-ended agent evolution from individual agents to groups that share and reuse experiences.
{\kg} first decomposes the idea into motivation, method, experiment, and impact claims, then projects the shared scholarly topology into a claim-level evidence bank, publication-grounded reviewer profiles, and idea-specific rubrics.
The evidence bank grounds individual claims in prior work, while the reviewer profiles and rubrics introduce the expert perspectives and evaluation requirements relevant to the proposal.
Together, these contexts condition reviewer-specific assessments representing evolution and open-ended AI, AI-agent evaluation, and machine learning and reinforcement learning.
The full meta-review consolidates reviewer consensus and disagreements into shared strengths and weaknesses, of which the figure presents only representative excerpts.
The displayed findings include agreement that the motivation addresses a limitation of tree-structured evolution, concerns about the underspecified shared-pool mechanism and incomplete failure-mode analysis, and differing interpretations of transferability and novelty.
These and additional findings are converted into revision advice, with the figure highlighting representative targets such as quantifying diversity utilization, specifying the shared-pool schema, auditing the evolution pipeline, strengthening statistical evidence, and validating transfer.
By projecting the same scholarly topology into idea-specific evidence, reviewer perspectives, and evaluation criteria, {\kg} standardizes the route from scientific claims to field-specific judgment and actionable revision.

\begin{figure*}[!th] 
  \vspace{-0.5cm}
  \centering
  \includegraphics[width=1\linewidth]{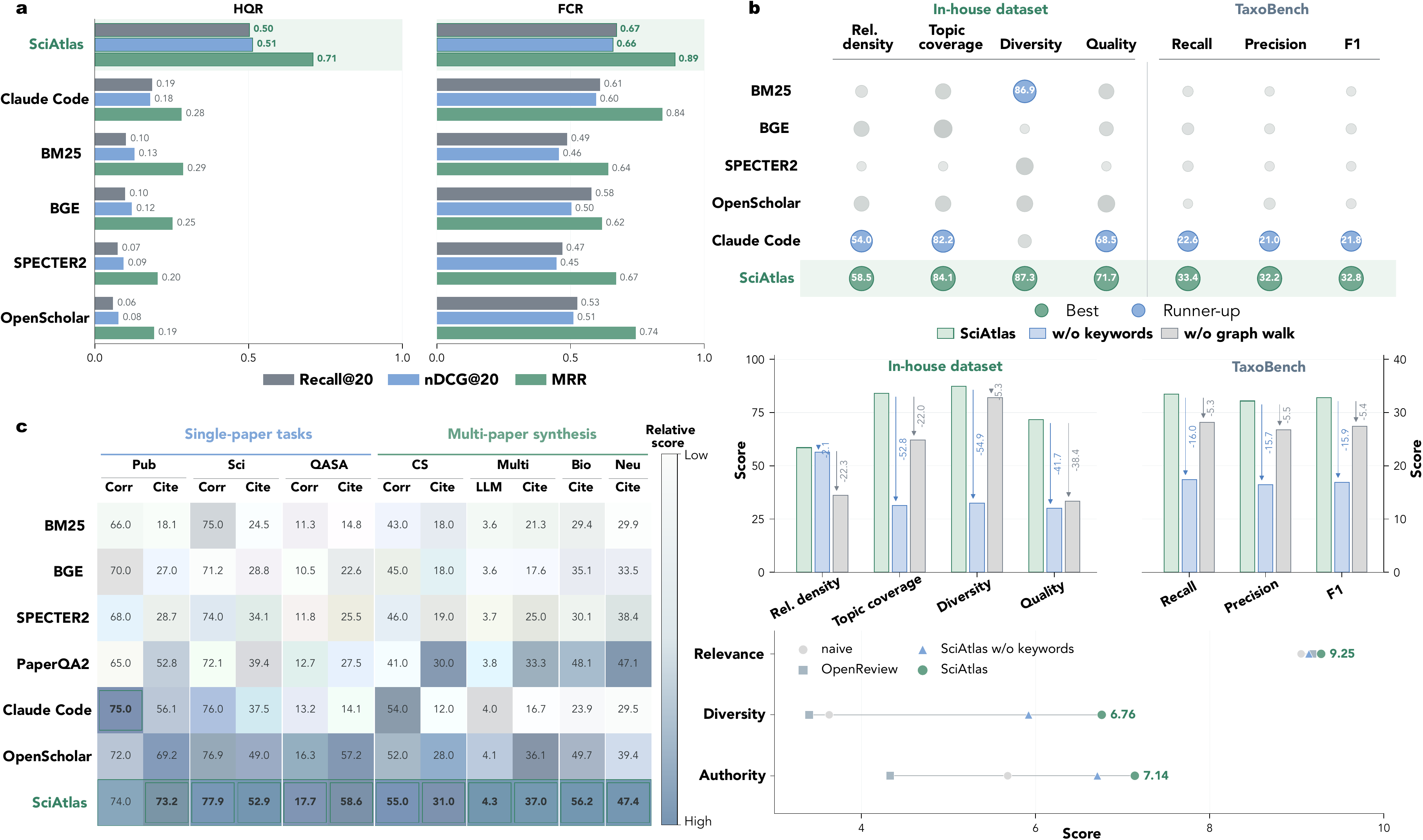}
  \caption{
    \small
    \textbf{Foundational evaluation of latent-relation recovery and scientific-object retrieval.}
    \textbf{a,} Latent relation recovery on \textsc{SciAtlasEval-HQR} and \textsc{SciAtlasEval-FCR}. Recall@20, nDCG@20, and mean reciprocal rank (MRR) evaluate whether each system can recover masked or temporally implicit scholarly connections from indirect evidence.
    \textbf{b,} Retrieval of core scholarly objects. Top, paper-retrieval performance on \textsc{SciAtlasEval-Idea} and TaxoBench; bubble size encodes metric value, and coloured markers denote the best and runner-up methods. Middle, component ablations of {\kg}, comparing the full setting with variants without keyword indexing or graph propagation; arrows indicate the absolute performance decrease from the full setting. Bottom, author-retrieval performance on relevance, diversity, and authority, comparing naive author aggregation, OpenReview, {\kg} without keywords, and the full {\kg} setting.
    \textbf{c,} Scientific question-answering and literature-synthesis performance on ScholarQABench. Results are reported for single-paper tasks (Pub, Sci and QASA) and multi-paper synthesis tasks (CS, Multi, Bio and Neu). Cells show correctness (Corr), citation quality (Cite) or LLM-judged answer quality (LLM), as applicable; outlined cells indicate the best-performing system within each metric. Higher values indicate better performance throughout.
}
  \vspace{-0.2cm}
  \label{fig:main_results}
\end{figure*}

\subsection{{\kg} provides a shared knowledge foundation across scientific workflows}

The preceding workflows require different scientific objects and relations, yet each relies on the same underlying process: grounding heterogeneous research inputs in a shared scholarly topology, propagating relevance across its relations, and projecting the resulting field into workflow-specific context.
To test whether their gains are supported by this common foundation, we evaluate its underlying capabilities independently of the three workflows.

\paragraph{Experimental setup}
We evaluate the shared foundation at three complementary levels: latent-relation recovery, scientific-object retrieval, and evidence-grounded synthesis.
For latent-relation recovery, \textsc{SciAtlasEval-HQR} comprises 200 heterogeneous query--object cases with direct query--target edges masked, whereas \textsc{SciAtlasEval-FCR} comprises 70 historical-paper cases that restrict retrieval to past literature and project the retrieved evidence to future papers through citations.
Both subsets are evaluated using Recall@20, nDCG@20, and MRR.
For scientific-object retrieval, we reuse \textsc{SciAtlasEval-Idea} to retrieve papers and authors from structured research ideas, evaluating paper sets by relevance density, topic coverage, diversity, and overall quality, and authors by relevance, expertise diversity, and authority.
We additionally use TaxoBench \cite{taxobench}, which contains 72 expert-authored survey taxonomies and 3,815 associated papers, and report recall, precision, and F1 against the expert-cited paper sets.
For evidence-grounded synthesis, ScholarQABench \cite{openscholar} provides 2,967 expert-written questions across computer science, physics, neuroscience and biomedicine, covering both single-paper question answering and multi-paper synthesis.
We report correctness, citation quality, and LLM-judged answer quality where applicable.
For latent-relation and paper-retrieval experiments, we compare {\kg} with five baselines: BM25, BGE \cite{bge}, SPECTER2 \cite{specter2}, OpenScholar \cite{openscholar} and Claude Code.
ScholarQABench additionally includes PaperQA2 \cite{paperqa2} and compares {\kg}-augmented OpenScholar with all six systems.
Author retrieval is compared with naive author aggregation and OpenReview expertise\footnote{\url{https://github.com/openreview/openreview-expertise}.}.
We further ablate the keyword layer and topological propagation to assess their contributions.
All configurable LLM components use \textsc{GPT-5.5}, and all methods receive the same queries and follow the same evaluation protocol within each benchmark.
Dataset construction, metric definitions, and baseline settings are detailed in Appx.~\ref{app:dataset}, \ref{app:metrics-hqr-fcr}, \ref{app:metrics-idea-paper}, \ref{app:metrics-idea-author}, \ref{app:metrics-taxobench}, \ref{app:metrics-scholarqabench} and \ref{app:baselines-foundational}.

\paragraph{Latent-relation recovery}
{\kg} achieves the strongest performance on both \textsc{SciAtlasEval-HQR} and \textsc{FCR} (Fig.~\ref{fig:main_results}a).
On HQR, its nDCG@20 reaches 0.51, nearly three times the 0.18 of the strongest baseline, while also leading Recall@20 and MRR.
The advantage persists under the temporal constraint of FCR, where {\kg} achieves the best results across all three metrics, including an nDCG@20 of 0.66 compared with 0.60 for the strongest baseline.
Because HQR masks direct query--target edges and FCR restricts retrieval to historical literature, these gains cannot rely on explicit connections or access to future evidence.
Instead, the shared schema grounds heterogeneous query objects in the same scholarly topology, allowing relevance to propagate through alternative citation, conceptual, and expertise paths.
The same process identifies historical neighbourhoods whose distributed signals anticipate connections that become explicit in later publications.
These results show that {\kg} recovers latent and temporally persistent scientific context through a common relational process, providing the structural foundation from which different workflows can subsequently retrieve their required objects and relations.

\paragraph{Scientific-object retrieval}
{\kg} also achieves the strongest paper-retrieval performance on both \textsc{SciAtlasEval-Idea} and TaxoBench (Fig.~\ref{fig:main_results}b).
On \textsc{SciAtlasEval-Idea}, it leads all four dimensions and reaches an overall quality score of 71.7, compared with 68.5 for the strongest baseline.
This advantage transfers to the external TaxoBench setting, where its F1 reaches 32.8 versus 21.8 for the strongest baseline.
The same retrieval process extends to authors: {\kg} maintains the highest relevance and authority while increasing expertise diversity to 6.76, compared with 3.63 for the strongest non-{\kg} method.
The ablations clarify the complementary roles of its two retrieval components.
Removing the keyword layer sharply reduces topic coverage, diversity, and overall quality, showing that fine-grained conceptual anchors connect scientifically related objects expressed through different terminology.
Disabling topological propagation instead produces the largest loss in relevance density and also substantially reduces overall quality, indicating that relational support is needed to distinguish structurally relevant objects from merely similar ones.
Rather than constructing separate acquisition pipelines, {\kg} grounds each input in the shared schema, propagates relevance across the scholarly topology, and projects the resulting field into paper or author space.
The required objects therefore change across tasks, while the process used to acquire and organize them remains consistent.

\paragraph{Evidence-grounded scientific synthesis}
Linking OpenScholar evidence passages through relations among their source papers in {\kg} consistently improves citation quality across all ScholarQABench tasks (Fig.~\ref{fig:main_results}c).
The largest gains appear in biomedicine, where citation quality increases from 49.7 to 56.2, and neuroscience, where it increases from 39.4 to 47.4.
{\kg}-augmented OpenScholar also improves correctness on QASA and Scholar-CS and LLM-judged answer quality on Scholar-Multi.
By contrast, the correctness gains on the Pub and Sci single-paper tasks remain comparatively small, defining where the scholarly topology contributes most.
When an answer depends mainly on evidence localized within one paper, additional relational context offers limited benefit; when evidence is distributed across multiple studies, relations among their source papers guide the retrieval of complementary support.
Because the answer-generation pipeline remains unchanged, these gains arise from transforming independent text passages into context organized by the shared scholarly topology.
Together with the relation- and object-level results, this evidence supports that the breadth, depth and standardization achieved across the preceding workflows derive from a common schema and retrieval mechanism rather than separate workflow-specific knowledge pipelines.

\section{Discussion}
This study establishes {\kg} as a reusable scholarly knowledge infrastructure for automated scientific research.
Its central contribution is not to provide every workflow or discipline with the same collection of scientific objects, because the evidence, relations, and expertise required by different research tasks necessarily vary.
Instead, {\kg} provides a common process for constructing these collections: heterogeneous research inputs are grounded in a shared schema, relevance is propagated across the scholarly topology, and the resulting relevance field is projected into the context required by each workflow.
The commonality therefore lies in the knowledge-acquisition pathway rather than in the retrieved content.
This distinction explains how the same infrastructure supports three different dimensions of scientific reasoning.
For research development trajectory reconstruction, it broadens the available context by organizing evidence into developmental stages, methodological branches, and representative research threads beyond locally prominent papers.
For research opportunity discovery, it deepens context construction by locating unresolved field-level bottlenecks and connecting them to structurally matched mechanisms within and beyond the immediate domain.
For innovation assessment, it standardizes context assembly by making relevant evidence, expert perspectives, and field-specific evaluation norms accessible through the same machine-actionable schema.
The foundational evaluations further show that these workflow-level gains rest on a common underlying capability.
Across relation recovery, scientific-object retrieval, and evidence-grounded synthesis, {\kg} consistently grounds heterogeneous inputs in the shared schema, propagates relevance beyond direct document similarity, and projects the resulting field into task-relevant context.
Component ablations further confirm the complementary roles of fine-grained conceptual anchoring and topological propagation.
These findings indicate that the improvements in breadth, depth, and standardization arise from a shared infrastructure for acquiring and organizing scientific context rather than from three independent workflow-specific pipelines.

Three limitations define the current scope of {\kg}.
First, although its graph-construction pipeline is automated, {\kg} is updated periodically rather than continuously, and its coverage depends on the availability of the underlying scholarly records.
This may delay the incorporation of emerging work and limit context recovery in rapidly evolving fields.
The availability of expertise and normative signals can also vary across disciplines and publication venues, limiting how completely the shared schema represents the communities and evaluation practices surrounding every research object.
Second, the present study focuses on knowledge-intensive stages of automated scientific research and does not examine closed-loop experimental validation.
Although {\kg} can retrieve experimental evidence and help formulate validation paths, experimental procedures, conditions, measurements, and outcomes are not yet represented as explicit objects that agents can directly access and reason over.
The present claims therefore concern scientific-context construction and knowledge-grounded decision support rather than autonomous empirical discovery.
Third, maintaining and querying a graph at the scale of {\kg} requires substantial storage, indexing and retrieval infrastructure.
The current implementation restricts propagation to query-centred local neighbourhoods to keep context construction computationally tractable, but lightweight or resource-constrained deployments may still require further systems optimization.

These limitations suggest three directions for extending {\kg}.
Continuous or incremental graph updates with explicit temporal provenance would allow agents to distinguish established structures from emerging signals while reducing delays in rapidly developing fields.
Broader integration of review records and other evaluative sources could further improve the coverage of field-specific norms and make their disciplinary boundaries more explicit.
An experimental layer could transform procedures, conditions, and validation evidence reported in papers into machine-actionable representations, linking scholarly context more directly to protocol design and empirical testing.
More efficient graph partitioning, caching and incremental indexing could reduce infrastructure overhead and support deployment under different computational and storage constraints.
Despite these opportunities for extension, the present results show that {\kg} already enables broader access, deeper relational context and more standardized knowledge assembly across distinct scientific workflows.
More broadly, by transforming scientific-context construction from a component repeatedly rebuilt within individual agents into shared infrastructure, {\kg} provides a persistent, machine-actionable knowledge foundation on which increasingly capable AI scientists can reason, collaborate and advance scientific research.

\section{Methods}

\subsection{Construction of {\kg}}

\paragraph{A shared graph schema for scientific knowledge infrastructure}
{\kg}'s shared schema integrates five complementary layers of scholarly knowledge (Fig.~\ref{fig:overview}b).
The \textbf{evidential layer} provides the paper-centred backbone through citation and publication relations.
We specifically construct the \textbf{conceptual layer} as a keyword graph comprising normalized keyword nodes, paper--keyword associations, and keyword co-occurrence relations to bridge implicit associations among scientific objects, connect equivalent or related concepts expressed through different terminology and levels of abstraction across disciplines, and provide fine-grained anchors for retrieval beyond direct document similarity.
The \textbf{disciplinary layer} situates papers within the hierarchical organization of science through topic, subfield, field, and domain nodes.
The \textbf{expertise layer} connects scholarly records to the researchers, institutions, and communities that produce them through authorship, affiliation, and co-authorship relations.
We further construct the \textbf{normative layer} by associating papers, where available, with peer-review and evaluation signals retained as paper-level attributes, making field-specific evaluation context accessible within the same schema.
Together, these layers place scientific evidence, concepts, disciplines, expertise, and evaluation signals within a single machine-readable scholarly topology, allowing different workflows to acquire and organize their required scientific objects through a common substrate.

\paragraph{Curation of scholarly records}
We curate raw scholarly records into graph-consistent entities and typed relations.
The primary source of {\kg} is OpenAlex\footnote{\url{https://openalex.org}.}, which provides large-scale metadata on papers, authors, institutions, venues, topics, references, citation statistics, and publication information.
We transform them into an infrastructure-oriented graph representation that can support multiple scientific workflows.
For each paper, we preserve attributes needed to situate it within the scholarly ecosystem, including title, abstract, venue, publication date, citation count, reference list, open-access status, and available full-text links.
For authors, institutions, and sources, we retain stable identifiers and descriptive metadata that support expertise modeling and research community retrieval.
We also preserve the hierarchical topic assignments provided for each paper to construct its links to topic, subfield, field, and domain nodes.
To populate the normative layer, we deploy an automated search agent that uses paper titles, authors, venues, and available identifiers to locate publicly accessible peer-review pages, retrieve their review comments and evaluation records, and associate them with the corresponding papers as structured attributes.
Before graph construction, we normalize paper titles, source names, and other textual fields to reduce formatting variation and duplicate surface forms.
Papers with missing core metadata, unavailable or extremely short abstracts, or unsupported languages are removed to ensure sufficient textual information for subsequent indexing and retrieval.
Entities are linked using stable identifiers whenever available, and normalized metadata are used for deduplication when identifiers are incomplete.
Ambiguous author records are handled conservatively to avoid merging distinct researchers with identical or similar names.
From the curated records, we construct citation relations between papers, authorship relations between papers and authors, affiliation relations between authors and institutions, publication relations between papers and sources, and hierarchical disciplinary relations connecting papers to topics, subfields, fields, and domains.
These entities, attributes, and typed relations form the evidential, disciplinary, expertise, and normative components of the shared scholarly topology.

\paragraph{A keyword-based conceptual backbone for bridging implicit scientific relations}
Citation links indicate explicit intellectual dependencies between papers, but do not reveal the concepts underlying these dependencies or fully connect scientifically related studies that do not connect directly.
So we construct a keyword-based conceptual layer to expose the conceptual basis of existing paper relations, align related concepts expressed through different terminology and levels of abstraction, and introduce implicit connections beyond the citation graph.
Specifically, we extract reusable scientific keywords from paper titles and abstracts, including mechanisms, materials, methods, datasets, tasks, diseases, models, and experimental settings.
Synonymous terms and formatting variants are mapped to normalized forms, with each normalized keyword represented as a unique conceptual node.
For every retained keyword in a paper, we construct a paper--keyword relation weighted by a salience score that records the importance of the keyword to the source paper.
We further connect each pair of keywords that occurs within the same paper and store its corpus-level co-occurrence frequency as an edge attribute.
Paper--keyword relations make explicit which concepts are shared by connected papers and provide interpretable conceptual paths between them.
Keyword normalization connects equivalent or related scientific objects expressed differently across studies and disciplines, while keyword co-occurrence introduces additional paths between concepts and papers whose relations remain implicit in the citation graph.
Together, these relations increase the relational coverage of the scholarly topology and provide fine-grained anchors and denser evidence paths for grounding heterogeneous inputs, propagating relevance and recovering latent scientific associations.

\paragraph{Neuro-symbolic indexing for heterogeneous graph grounding}
To make the shared scholarly topology accessible from heterogeneous research inputs, we construct complementary symbolic and neural indexes over its scientific objects.
The symbolic indexes map normalized identifiers, canonical names, and surface forms to their corresponding nodes, covering paper titles, keyword strings, author names, institution names, publication sources, and disciplinary labels.
They provide high-precision resolution when an input explicitly names a paper, concept, researcher, institution, venue, or research field.
In parallel, we encode retrieval-relevant textual fields, including paper titles and abstracts, keyword descriptions, and disciplinary-topic descriptions, into dense representations and organize them using approximate nearest-neighbour indexes.
These neural indexes map broader or underspecified inputs to semantically related regions of the graph even when the input and indexed objects do not share the same wording.
Given a research input, symbolic retrieval preserves precise links to explicitly mentioned entities, while neural retrieval supplies semantically relevant candidates for objects that cannot be resolved exactly.
The retrieved heterogeneous nodes form the candidate seeds for subsequent query-to-graph anchoring and topological propagation.

\subsection{Unified neuro-symbolic retrieval on {\kg}}
\label{sec:retrieval}

Given a heterogeneous research query, {\kg} applies a unified neuro-symbolic retrieval mechanism over the shared scholarly topology (Fig.~\ref{fig:overview}c).
It grounds the input to typed graph nodes, propagates relevance through a query-centred and relation-aware topology, and projects the resulting relevance field into workflow-specific scientific objects.
So the required objects and relations can vary across tasks and disciplines while their acquisition follows the same process.

\subsubsection{Heterogeneous query-to-graph anchoring}

Given an input $\boldsymbol{x}$, an LLM-based parser decomposes it into typed query units:
\begin{align}
\boldsymbol{\mathcal{U}}(\boldsymbol{x})=\{(\boldsymbol{u}_i,\boldsymbol{\tau}_i,\boldsymbol{a}_i)\}_{i=1}^{m},
\end{align}
where $\boldsymbol{u}_i$ is an extracted mention or semantic unit, $\boldsymbol{\tau}_i$ specifies its scholarly object type, such as concept, paper, author, or topic, and $\boldsymbol{a}_i\in[0,1]$ denotes its salience.
The constrained output schema routes each unit to the corresponding anchoring channel, converting heterogeneous inputs into typed graph signals.

Concept units are anchored to normalized keyword nodes through exact and embedding-based matching:
\begin{align}
\boldsymbol{s}^{\mathrm{con}}_g
=
\max_{i:\boldsymbol{\tau}_i=\mathrm{concept}}
\max\left(
\mathbbm{I}[\boldsymbol{u}_i=\boldsymbol{g}]\cdot \boldsymbol{a}_i,\;
\mathbbm{I}[\mathrm{sim}(\boldsymbol{u}_i,\boldsymbol{g})\geq \theta_{\mathrm{con}}]\cdot \boldsymbol{a}_i\cdot \mathrm{sim}(\boldsymbol{u}_i,\boldsymbol{g})
\right),
\end{align}
where exact equality is evaluated after keyword normalization, $\mathrm{sim}(\cdot,\cdot)$ denotes cosine similarity between dense representations, and $\theta_{\mathrm{con}}$ is the semantic matching threshold.
The score retains the strongest exact or semantic match assigned to keyword node $\boldsymbol{g}$ across all concept units.
This channel maps explicit concepts and terminology variants to shared conceptual anchors, from which relevance can propagate through paper--keyword and keyword co-occurrence relations.

Document units are anchored to paper nodes by combining title-level resolution with abstract-level semantic matching:
\begin{align}
\boldsymbol{s}^{\mathrm{doc}}_p
=
\frac{
\boldsymbol{\lambda}_{\mathrm{title}}\boldsymbol{s}^{\mathrm{title}}_p
+
\boldsymbol{\lambda}_{\mathrm{abs}}\boldsymbol{s}^{\mathrm{abs}}_p
}{
\boldsymbol{\lambda}_{\mathrm{title}}\mathbbm{I}[\boldsymbol{s}^{\mathrm{title}}_p \text{ exists}]
+
\boldsymbol{\lambda}_{\mathrm{abs}}\mathbbm{I}[\boldsymbol{s}^{\mathrm{abs}}_p \text{ exists}]
},
\qquad
\begin{aligned}
\boldsymbol{s}^{\mathrm{title}}_p
&=
\max_{i:\boldsymbol{\tau}_i\in\{\mathrm{title},\mathrm{reference}\}}
\boldsymbol{a}_i\cdot f_{\mathrm{title}}(\boldsymbol{u}_i,\mathrm{title}(\boldsymbol{p})),\\
\boldsymbol{s}^{\mathrm{abs}}_p
&=
\max_{i:\boldsymbol{\tau}_i\in\{\mathrm{abstract},\mathrm{claim},\mathrm{idea},\mathrm{paper}\}}
\boldsymbol{a}_i\cdot \mathrm{sim}(\boldsymbol{u}_i,\mathrm{abs}(\boldsymbol{p})).
\end{aligned}
\end{align}
Here $f_{\mathrm{title}}$ combines normalized exact matching, fuzzy sequence similarity and token overlap, whereas $\mathrm{sim}$ measures semantic similarity between a paper-like unit and the abstract of $\boldsymbol{p}$.
The weighted mean is computed over available components, and $\boldsymbol{s}^{\mathrm{doc}}_p$ is set to zero when neither component exists.

Author, institution, venue, and disciplinary-topic units are anchored to their corresponding node types:
\begin{align}
\boldsymbol{s}^{\mathrm{ent}}_v
=
\max_{i:\boldsymbol{\tau}_i=\mathrm{type}(\boldsymbol{v})}
\boldsymbol{a}_i\cdot f_{\mathrm{ent}}(\boldsymbol{u}_i,\boldsymbol{v}),
\end{align}
where $f_{\mathrm{ent}}$ combines normalized exact matching, alias matching, and fuzzy token overlap.
This channel preserves high-precision graph entry points for explicitly named scholarly entities.

In parallel, cross-type semantic anchoring uses the representation of the complete input to retrieve globally relevant nodes.
For each indexed node $\boldsymbol{v}$ of type $t\in\boldsymbol{\mathcal{T}}$, such as a paper, keyword or topic, we compute:
\begin{align}
\boldsymbol{s}^{\mathrm{sem}}_v
=
\mathbbm{I}[\mathrm{sim}(\mathbf{e}_x,\mathbf{e}_v)\geq \theta_t]
\cdot
\boldsymbol{\lambda}_t
\cdot
\mathrm{sim}(\mathbf{e}_x,\mathbf{e}_v),
\end{align}
where $\mathbf{e}_x$ and $\mathbf{e}_v$ are the dense representations of $\boldsymbol{x}$ and $\boldsymbol{v}$, respectively, $\theta_t$ is a type-specific matching threshold, and $\boldsymbol{\lambda}_t$ weights the contribution of node type $t$.
Unlike concept anchoring, which preserves the fine-grained structure of extracted concept units, this channel provides a global entry point for broad or underspecified inputs.

We combine the anchoring channels within the shared node space and define the heterogeneous seed set as:
\begin{align}
\boldsymbol{\mathcal{S}}(\boldsymbol{x})
=
\{(\boldsymbol{v},\boldsymbol{s}^{\mathrm{anc}}_v)\mid \boldsymbol{s}^{\mathrm{anc}}_v>0\},\quad
\boldsymbol{s}^{\mathrm{anc}}_v
=
\max\left(
\boldsymbol{s}^{\mathrm{con}}_v,
\boldsymbol{s}^{\mathrm{doc}}_v,
\boldsymbol{s}^{\mathrm{ent}}_v,
\boldsymbol{s}^{\mathrm{sem}}_v
\right),
\end{align}
where scores from channels not applicable to node $\boldsymbol{v}$ are set to zero.
We min--max normalize $\boldsymbol{s}^{\mathrm{anc}}_v$ to $[0,1]$ within each node type and retain the same notation for the normalized score.
The resulting typed seeds vary with the input and target task while providing a common representation for query-centred topology construction and relevance propagation.

\subsubsection{Query-centred topology construction and weighting}

The typed seeds define where the input enters {\kg}.
We construct a query-centred local topology by expanding deterministically from $\boldsymbol{\mathcal{S}}(\boldsymbol{x})$ along typed graph relations for a fixed number of hops:
\begin{align}
\boldsymbol{\mathcal{G}}_{\boldsymbol{x}}
=
(\boldsymbol{\mathcal{V}}_{\boldsymbol{x}},\boldsymbol{\mathcal{E}}_{\boldsymbol{x}}).
\end{align}
We limit the neighbours retained for each node type to keep the topology tractable and prevent high-degree nodes from dominating it before relevance is estimated.
Within $\boldsymbol{\mathcal{G}}_{\boldsymbol{x}}$, each seed receives an unnormalized weight combining its anchoring strength, node-type prior, and scholarly importance:
\begin{align}
\boldsymbol{w}^{\mathrm{seed}}_v
=
\boldsymbol{\mu}_{\mathrm{type}(\boldsymbol{v})}
\cdot
\boldsymbol{s}^{\mathrm{anc}}_v
\cdot
\left(1+\boldsymbol{\gamma}\cdot \boldsymbol{\phi}_{\boldsymbol{x}}(\boldsymbol{v})\right),
\end{align}
where $\boldsymbol{\mu}_{\mathrm{type}(\boldsymbol{v})}$ balances different seed types, $\boldsymbol{\gamma}$ controls the importance contribution, and $\boldsymbol{\phi}_{\boldsymbol{x}}(\boldsymbol{v})$ is normalized within the query-centred context.
For papers, we derive this importance from citation counts normalized across local paper nodes:
\begin{align}
\label{eq:importance_paper}
\boldsymbol{\phi}_{\boldsymbol{x}}(\boldsymbol{p})
=
\min\left(
1,
\frac{\log(1+\boldsymbol{c}_p)}
{\log(1+\max_{\boldsymbol{p}'\in\boldsymbol{\mathcal{P}}_{\boldsymbol{x}}}\boldsymbol{c}_{p'})}
\right),
\end{align}
where $\boldsymbol{c}_p$ is the citation count of $\boldsymbol{p}$ and $\boldsymbol{\mathcal{P}}_{\boldsymbol{x}}$ is the set of local paper nodes.
For authors, we use the mean citation count of their publications, normalized across local author nodes:
\begin{align}
\label{eq:importance_author}
\boldsymbol{\phi}_{\boldsymbol{x}}(\boldsymbol{a})
=
\min\left(
1,
\frac{
\log\left(1+
\frac{1}{|\boldsymbol{\mathcal{P}}_a|}
\sum_{\boldsymbol{p}\in\boldsymbol{\mathcal{P}}_a}
\boldsymbol{c}_p
\right)
}{
\log\left(1+
\max_{\boldsymbol{a}'\in\boldsymbol{\mathcal{A}}_{\boldsymbol{x}}}
\frac{1}{|\boldsymbol{\mathcal{P}}_{a'}|}
\sum_{\boldsymbol{p}\in\boldsymbol{\mathcal{P}}_{a'}}
\boldsymbol{c}_p
\right)
}
\right),
\end{align}
where $\boldsymbol{\mathcal{P}}_a$ contains papers authored by $\boldsymbol{a}$ and $\boldsymbol{\mathcal{A}}_{\boldsymbol{x}}$ contains local author nodes.
We set $\boldsymbol{\phi}_{\boldsymbol{x}}(\boldsymbol{v})=0$ for node types without an importance prior and normalize the seed weights into the restart distribution:
\begin{align}
\boldsymbol{r}^{(0)}_v
=
\boldsymbol{q}_v
=
\begin{cases}
\frac{\boldsymbol{w}^{\mathrm{seed}}_v}{\sum_{\boldsymbol{u}\in\boldsymbol{\mathcal{S}}(\boldsymbol{x})}\boldsymbol{w}^{\mathrm{seed}}_u},
& \boldsymbol{v}\in\boldsymbol{\mathcal{S}}(\boldsymbol{x}),\\
0,
& \boldsymbol{v}\notin\boldsymbol{\mathcal{S}}(\boldsymbol{x}).
\end{cases}
\end{align}

We preserve the semantics of the shared schema by assigning each edge a relation-aware propagation weight:
\begin{align}
\boldsymbol{w}(\boldsymbol{u},\boldsymbol{v},\boldsymbol{\ell})
=
\boldsymbol{\beta}_{\boldsymbol{\ell}}
\cdot
\boldsymbol{\eta}(\boldsymbol{u},\boldsymbol{v},\boldsymbol{\ell}),
\end{align}
where $\boldsymbol{\beta}_{\boldsymbol{\ell}}$ controls the contribution of relation type $\boldsymbol{\ell}$ and $\boldsymbol{\eta}$ incorporates relation-specific attributes.
The base weights distinguish citation, paper--keyword, keyword co-occurrence, topic-assignment, authorship, affiliation, and publication relations instead of treating the local topology as homogeneous.
For a paper--keyword edge, the modifier combines keyword salience with its anchoring to the query:
\begin{align}
\boldsymbol{\eta}(\boldsymbol{p},\boldsymbol{g},\mathrm{has\_keyword})
=
\boldsymbol{\mathrm{rel}}_{\boldsymbol{p},\boldsymbol{g}}
\cdot
\begin{cases}
1+\boldsymbol{s}^{\mathrm{anc}}_g,
& \boldsymbol{g}\in\boldsymbol{\mathcal{S}}(\boldsymbol{x}),\\
\boldsymbol{\epsilon}_{\mathrm{kw}},
& \boldsymbol{g}\notin\boldsymbol{\mathcal{S}}(\boldsymbol{x}),
\end{cases}
\end{align}
where $\boldsymbol{\mathrm{rel}}_{\boldsymbol{p},\boldsymbol{g}}$ is the keyword salience stored on the edge and $\boldsymbol{\epsilon}_{\mathrm{kw}}$ retains a smoothed path through non-seed keywords.
For frequency-based relations, including keyword co-occurrence and co-authorship, we use
\begin{align}
\boldsymbol{\eta}(\boldsymbol{u},\boldsymbol{v},\boldsymbol{\ell})
=
\max\left(
1,
\min\left(
\boldsymbol{c}_{\max},
\log(1+\boldsymbol{n}_{\boldsymbol{u},\boldsymbol{v}})
\right)
\right),
\end{align}
where $\boldsymbol{n}_{\boldsymbol{u},\boldsymbol{v}}$ is the observed frequency and $\boldsymbol{c}_{\max}$ limits the influence of highly frequent relations.
For other relations, we set $\boldsymbol{\eta}=1$.
These modifiers convert the conceptual layer into query-sensitive paths while retaining weaker implicit connections for subsequent propagation.
The normalized seed weights and relation-aware edges together define the query-centred weighted topology used for restarted propagation.

\subsubsection{Restarted topological propagation}

On the weighted topology $\boldsymbol{\mathcal{G}}_{\boldsymbol{x}}$, let $\boldsymbol{\mathcal{N}}(\boldsymbol{u})$ denote the neighbours of node $\boldsymbol{u}$ and $\boldsymbol{\ell}_{uv}$ the relation type connecting $\boldsymbol{u}$ and $\boldsymbol{v}$.
We normalize the relation-aware edge weights into transition probabilities:
\begin{align}
\boldsymbol{P}(\boldsymbol{v}\mid \boldsymbol{u})
=
\frac{
\boldsymbol{w}(\boldsymbol{u},\boldsymbol{v},\boldsymbol{\ell}_{uv})
}{
\sum_{\boldsymbol{z}\in\boldsymbol{\mathcal{N}}(\boldsymbol{u})}
\boldsymbol{w}(\boldsymbol{u},\boldsymbol{z},\boldsymbol{\ell}_{uz})
}.
\end{align}
We set $\boldsymbol{P}(\boldsymbol{v}\mid\boldsymbol{u})=0$ for non-neighbouring nodes and $\boldsymbol{P}(\boldsymbol{u}\mid\boldsymbol{u})=1$ when $\boldsymbol{u}$ has no neighbours.
The shared operator traverses graph relations bidirectionally, whereas ROD uses a separately constructed directed citation-evolution graph to preserve how later studies develop from earlier work.

Let $\boldsymbol{r}^{(t)}_v$ denote the relevance score of node $\boldsymbol{v}$ at iteration $t$. We initialize the propagation with the restart distribution $\boldsymbol{r}^{(0)}_v=\boldsymbol{q}_v$.
At each iteration, the score of each node is updated by combining restart to the original seed distribution with diffusion from neighbouring nodes:
\begin{align}
\boldsymbol{r}^{(t+1)}_v
=
\alpha \boldsymbol{q}_v
+
(1-\alpha)
\sum_{\boldsymbol{u}\in\boldsymbol{\mathcal{V}}_{\boldsymbol{x}}}
\boldsymbol{r}^{(t)}_u
\boldsymbol{P}(\boldsymbol{v}\mid \boldsymbol{u}),
\end{align}
where $\alpha$ is the restart probability. The restart term preserves the query-specific anchoring signal, whereas the diffusion term accumulates support from the surrounding scholarly topology.
If a node has no neighbours, its probability mass is retained on itself.
The propagation stops when the score vector converges or reaches the maximum number of iterations:
\begin{align}
\left\|
\boldsymbol{r}^{(t+1)}
-
\boldsymbol{r}^{(t)}
\right\|_1
<
\boldsymbol{\epsilon}
\quad
\text{or}
\quad
t=\boldsymbol{T}_{\max}.
\end{align}
The final score of each node $\boldsymbol{v}$ is denoted by $\boldsymbol{r}_v=\boldsymbol{r}^{(t^\star)}_v$, where $t^\star$ is the stopping iteration.
Together, these scores define a shared relevance field over the query-centred scholarly topology.
Then the projection operators translate this node-level field into scholarly-object pools that can be assembled into the context required by different scientific workflows.

\subsubsection{Scientific-object projection from the shared relevance field}

The shared relevance field spans heterogeneous node types, whereas each scientific workflow requires particular objects for context assembly.
For the workflows studied here, we project this field into ranked paper and author pools, which provide evidence and expertise objects, respectively.
The projected objects vary with the input and workflow, while their acquisition follows the same grounding, propagation, and projection process.

For each local paper node $\boldsymbol{p}\in\boldsymbol{\mathcal{P}}_{\boldsymbol{x}}$, we combine propagated relevance, direct anchoring, and context-relative importance:
\begin{align}
\boldsymbol{s}^{\mathrm{paper}}_p
=
\boldsymbol{\lambda}_{\mathrm{graph}}\boldsymbol{r}_p
+
\boldsymbol{\lambda}_{\mathrm{anc}}\boldsymbol{s}^{\mathrm{anc}}_p
+
\boldsymbol{\lambda}_{\mathrm{imp}}\boldsymbol{\phi}_{\boldsymbol{x}}(\boldsymbol{p}),
\end{align}
where $\boldsymbol{r}_p$ is its propagated score, $\boldsymbol{s}^{\mathrm{anc}}_p$ its normalized anchoring score, and $\boldsymbol{\phi}_{\boldsymbol{x}}(\boldsymbol{p})$ its importance defined in Eq.~\ref{eq:importance_paper}.
We set $\boldsymbol{s}^{\mathrm{anc}}_p=0$ for papers reached only through propagation and select the highest-scoring candidates:
\begin{align}
\boldsymbol{\mathcal{P}}_{\mathrm{rec}}(\boldsymbol{x})
=
\mathrm{TopK}_{\boldsymbol{p}\in\boldsymbol{\mathcal{P}}_{\boldsymbol{x}}}
\ \boldsymbol{s}^{\mathrm{paper}}_p .
\end{align}
This projection places directly matched and topologically recovered evidence within the same ranked paper pool.

For each local author node $\boldsymbol{a}\in\boldsymbol{\mathcal{A}}_{\boldsymbol{x}}$, we combine propagated relevance, direct anchoring, the relevance of the author's local publications, and author-level importance:
\begin{align}
\boldsymbol{s}^{\mathrm{author}}_a
=
\boldsymbol{\lambda}_{\mathrm{graph}}\boldsymbol{r}_a
+
\boldsymbol{\lambda}_{\mathrm{anc}}\boldsymbol{s}^{\mathrm{anc}}_a
+
\boldsymbol{\lambda}_{\mathrm{pub}}
\max_{\boldsymbol{p}\in\boldsymbol{\mathcal{P}}_a\cap\boldsymbol{\mathcal{P}}_{\boldsymbol{x}}}
\boldsymbol{s}^{\mathrm{paper}}_p
+
\boldsymbol{\lambda}_{\mathrm{imp}}\boldsymbol{\phi}_{\boldsymbol{x}}(\boldsymbol{a}),
\end{align}
where $\boldsymbol{\mathcal{P}}_a$ contains papers authored by $\boldsymbol{a}$ and $\boldsymbol{\phi}_{\boldsymbol{x}}(\boldsymbol{a})$ is defined in Eq.~\ref{eq:importance_author}.
We set the anchoring term to zero for authors reached only through propagation and the publication term to zero when $\boldsymbol{\mathcal{P}}_a\cap\boldsymbol{\mathcal{P}}_{\boldsymbol{x}}=\varnothing$.
The ranked author pool is
\begin{align}
\boldsymbol{\mathcal{A}}_{\mathrm{rec}}(\boldsymbol{x})
=
\mathrm{TopK}_{\boldsymbol{a}\in\boldsymbol{\mathcal{A}}_{\boldsymbol{x}}}
\ \boldsymbol{s}^{\mathrm{author}}_a .
\end{align}
RDT and ROD primarily use the projected paper pool, whereas IA additionally uses the author pool to assemble evidence-grounded expert perspectives.

\subsection{Workflow-specific scientific context assembly}

The shared retrieval mechanism provides a common process for acquiring scientific objects, while each workflow organizes the projected objects into context suited to its scientific objective (Fig.~\ref{fig:overview}d).
We define three context-assembly procedures for research development trajectory, research opportunity discovery, and scientific innovation assessment.

\subsubsection{Research development trajectory}
\label{sec:rdt}
Research Development Trajectory (RDT) organizes paper evidence projected from the shared relevance field into development stages, research threads, and evidence roles.
Given a topic, research question, or paper-centred description $\boldsymbol{x}$, it iteratively diagnoses and fills coverage gaps across these dimensions to reconstruct how a field emerges, branches, and advances.

\paragraph{Topic profiling and temporal scaffolding}
An LLM agent first constructs a topic profile specifying the normalized scope of $\boldsymbol{x}$, candidate terminology, potential research threads, and exclusion boundaries.
It then formulates a broad probe query $\boldsymbol{x}_{\mathrm{probe}}$ and applies the shared paper projection to obtain the initial pool $\boldsymbol{\mathcal{P}}^{(0)}=\boldsymbol{\mathcal{P}}_{\mathrm{rec}}(\boldsymbol{x}_{\mathrm{probe}})$.
The probe evidence resolves alternative expressions and ambiguous interpretations of the topic while estimating its temporal extent.
From the topic profile and publication-time distribution, a planning agent proposes ordered topic-dependent time windows:
\begin{align}
\boldsymbol{\mathcal{W}}_{\boldsymbol{x}}
=
\left\{
\boldsymbol{W}_b
=
\left[
t_b^{\mathrm{start}},
t_b^{\mathrm{end}}
\right]
\right\}_{b=1}^{\boldsymbol{B}},
\end{align}
where $\boldsymbol{\mathcal{W}}_{\boldsymbol{x}}$ contains $\boldsymbol{B}$ potentially unequal development stages.
Their boundaries follow the temporal extent and development pace observed in the probe evidence, preserving distinct foundational and recent-frontier periods.

\paragraph{Coverage-guided multi-round retrieval}
A single retrieval query may favour recent, lexically explicit, or prominent papers while overlooking earlier stages, less visible research threads, and specialized evidence roles.
We therefore formulate evidence acquisition as a coverage-guided iterative process.
At round $h$, an agent constructs a set of search actions:
\begin{align}
\boldsymbol{\mathcal{A}}^{(h)}
=
\left\{
\boldsymbol{a}^{(h)}_j
\right\}_{j=1}^{J_h},
\end{align}
where each action $\boldsymbol{a}^{(h)}_j$ specifies a subquery $\boldsymbol{x}^{(h)}_j$, target research thread, time window and evidence role.
Each subquery follows the shared anchoring, propagation, and paper-projection procedure in Sec.~\ref{sec:retrieval}, and the retrieved papers are merged with the accumulated evidence:
\begin{align}
\boldsymbol{\mathcal{P}}^{(h)}
=
\operatorname{Dedup}
\left(
\boldsymbol{\mathcal{P}}^{(h-1)}
\cup
\bigcup_{j=1}^{J_h}
\boldsymbol{\mathcal{P}}_{\mathrm{rec}}
\left(\boldsymbol{x}^{(h)}_j\right)
\right).
\end{align}

After each round, papers are organized by research thread, time window, and evidence role.
Research threads group papers by their methodological approaches to the focal problem, while evidence roles distinguish foundational studies, original methods, representative applications, benchmarks or datasets, surveys, and frontier work.
Papers with uncertain assignments remain as peripheral evidence.
A diagnosis agent then identifies gaps in these three coverage dimensions and records their targets, severity, query seeds, and whether they block termination.
We denote the resulting diagnosis by $\boldsymbol{\mathcal{D}}^{(h)}$ and its high-severity blocking gaps by $\boldsymbol{\mathcal{D}}_{\mathrm{high}}^{(h)}$.
These gaps are converted into targeted actions for round $h+1$, allowing the direction and depth of retrieval to adapt to the evidence missing from the current reconstruction.

\paragraph{Adaptive termination and trajectory assembly}
The retrieval loop terminates when no high-priority blocking gap remains, or the maximum number of rounds is reached:
\begin{align}
h^{\star}
=
\min
\left\{
h:
\boldsymbol{\mathcal{D}}_{\mathrm{high}}^{(h)}
=
\varnothing
\ \text{or}\
h=H_{\max}
\right\},
\end{align}
where $H_{\max}$ is the maximum permitted retrieval round.

After termination, we recluster $\boldsymbol{\mathcal{P}}^{(h^\star)}$ because refinement may reveal new research threads or alter existing cluster boundaries.
The papers are organized into a method--time evidence map that records when each thread emerges, expands, or declines.
Representative papers are selected using their graph-derived scores and coverage of research threads, development stages, and evidence roles.
The resulting trajectory is
\begin{align}
\boldsymbol{\mathcal{J}}(\boldsymbol{x})
=
\left\{
\left(
\boldsymbol{W}_b,
\boldsymbol{\mathcal{C}}_b,
\boldsymbol{\mathcal{P}}^{\mathrm{rep}}_b
\right)
\right\}_{b=1}^{\boldsymbol{B}},
\end{align}
where $\boldsymbol{\mathcal{C}}_b$ contains the research threads active in stage $\boldsymbol{W}_b$ and $\boldsymbol{\mathcal{P}}^{\mathrm{rep}}_b$ their representative papers.
Transitions between adjacent stages are summarized through methodological shifts, emerging concepts, and unresolved debates.

The structured trajectory is converted into a review outline whose sections correspond to particular development stages and research threads.
Each section receives an evidence pack containing its assigned papers, constraining generation to the evidence supporting that part of the trajectory.
RDT thereby assembles coverage across periods, branches, and evidence roles into a structured account of field development.

\subsubsection{Research opportunity discovery}
\label{sec:opportunity}
Given a research topic, seed paper, preliminary idea, or paper-centred description $\boldsymbol{x}$, Research Opportunity Discovery (ROD) organizes projected paper evidence into a local citation-evolution graph.
It abstracts field-level bottlenecks into a Research Structure Signature, retrieves structurally matched mechanisms from same-field and cross-domain neighbourhoods, and maps them into testable research opportunities.

\paragraph{Research graph construction}
An LLM agent first refines $\boldsymbol{x}$ into a focal research problem, technical keywords, and a retrieval-oriented question, from which it generates complementary seed queries covering distinct directions of the problem.
Each query is processed through the shared paper projection, and the candidates are reranked by substantive relevance and directional diversity to form the seed set $\boldsymbol{\mathcal{P}}_{\mathrm{seed}}$ while excluding weakly related and redundant papers.
Starting from these seeds, {\kg} expands along directed citation relations to predecessor and subsequent work while preserving temporal consistency.
The resulting citation-evolution graph is:
\begin{align}
\boldsymbol{\mathcal{G}}_{\mathrm{opp}}(\boldsymbol{x})
=
\left(
\boldsymbol{\mathcal{P}}_{\mathrm{opp}},
\boldsymbol{\mathcal{E}}_{\mathrm{cite}},
\boldsymbol{\mathcal{E}}_{\mathrm{kw}},
\boldsymbol{\mathcal{C}}_{\mathrm{phase}}
\right),
\end{align}
where $\boldsymbol{\mathcal{P}}_{\mathrm{opp}}$ is the local paper set, $\boldsymbol{\mathcal{E}}_{\mathrm{cite}}$ contains directed citation-evolution relations, $\boldsymbol{\mathcal{E}}_{\mathrm{kw}}$ contains paper--keyword links and $\boldsymbol{\mathcal{C}}_{\mathrm{phase}}$ contains temporally ordered research phases.
The citation relations trace how later studies develop from earlier work, while the keyword links expose recurring concepts and potential bridges across research branches.

\paragraph{Research structure signature extraction}
An LLM agent analyzes predecessor chains, parallel branches, research phases, and recurring limitations in $\boldsymbol{\mathcal{G}}_{\mathrm{opp}}(\boldsymbol{x})$ to identify unresolved scientific bottlenecks supported across the local field structure.
It excludes deficiencies of the graph itself, such as missing papers or incomplete connections, from consideration as research gaps.
The resulting Research Structure Signature $\boldsymbol{\Omega}_{\boldsymbol{x}}$ records the target domain, core gap, key objects and relations, unresolved tensions, desired solution properties, possible mechanism types, and inspiration-search hints.
Rather than proposing an idea, $\boldsymbol{\Omega}_{\boldsymbol{x}}$ preserves the constraints of the target problem while expressing its bottleneck in a mechanism-oriented form that guides structurally matched retrieval.

\paragraph{Multi-radius mechanism retrieval}
Guided by $\boldsymbol{\Omega}_{\boldsymbol{x}}$ and the field trend, a radius-gating agent selects same-field retrieval, cross-domain retrieval, or both, together with the number of candidates obtained from each radius.
Same-field retrieval prioritizes mechanisms compatible with the target objects, assumptions, and evaluation practices, whereas cross-domain retrieval expands the solution space when the bottleneck calls for mechanism transfer.
For the latter, the agent reformulates the core gap in terms of mechanisms, constraints, and relational patterns rather than repeating its domain-specific terminology.
At feedback round $r$, the resulting action set $\boldsymbol{\mathcal{A}}_{\mathrm{opp}}^{(r)}$ contains mechanism-oriented queries, target concepts, and the intended role of each inspiration.
Each query is processed through the shared paper projection, and the retrieved candidates are combined as:
\begin{align}
\boldsymbol{\mathcal{I}}^{(r)}
=
\operatorname{Dedup}
\left(
\bigcup_{\boldsymbol{a}^{(r)}_j\in\boldsymbol{\mathcal{A}}_{\mathrm{opp}}^{(r)}}
\boldsymbol{\mathcal{P}}_{\mathrm{rec}}
\left(
\boldsymbol{x}^{(r)}_j
\right)
\right),
\end{align}
where $\boldsymbol{x}^{(r)}_j$ is the query associated with action $\boldsymbol{a}^{(r)}_j$ and $\boldsymbol{\mathcal{I}}^{(r)}$ is the candidate inspiration pool.
Conceptual anchoring and topological retrieval allow this pool to include mechanisms connected through implicit conceptual relations rather than surface terminology alone.

\paragraph{Mechanism mapping and opportunity generation}
Each candidate in $\boldsymbol{\mathcal{I}}^{(r)}$ is assessed for a structural or mechanistic correspondence with the bottleneck represented by $\boldsymbol{\Omega}_{\boldsymbol{x}}$.
Candidates based only on surface vocabulary are removed, and the retained set $\boldsymbol{\mathcal{I}}_{\mathrm{use}}^{(r)}$ records the source mechanism, target object or bottleneck, adaptation, and transfer rationale.
The agent then generates $\boldsymbol{o}^{(r)}$ from the original input, citation-evolution graph, Research Structure Signature, retained mechanisms, and previous feedback.
Its structured output contains a graph-grounded motivation, proposed mechanism, traceable analogy mapping, primary falsifiable hypothesis, and phased experimental plan.
The initial validation isolates the claimed mechanism by varying one primary factor while holding potential confounders fixed, and specifies concrete baselines or controls, mechanism-matched ablations where applicable, and measurements that rule out alternative explanations.
Secondary variables, broader scope, and performance optimization are introduced only after the primary mechanism is supported.

\paragraph{Creativity--feasibility feedback and final opportunity brief}
At round $r$, a feedback agent evaluates $\boldsymbol{o}^{(r)}$ along creativity and feasibility.
Creativity measures whether the opportunity introduces a distinct mechanism, relation or hypothesis beyond the closest existing work, whereas feasibility examines its compatibility with the target objects, controllable variables, available experimental or computational settings, and validation route.
The resulting feedback $\boldsymbol{f}^{(r)}$ records the closest prior work, novelty risks, mechanism-transfer plausibility, experimental weaknesses, and required revisions.
Insufficient creativity redirects the next round toward more distant or structurally different mechanisms, whereas insufficient feasibility favours closer-field mechanisms, stronger constraints or a more controlled validation plan.
The loop terminates when both criteria reach their acceptance thresholds or the maximum number of feedback rounds is reached.
The retained output is
\begin{align}
\boldsymbol{\mathcal{O}}(\boldsymbol{x})
=
\left(
\boldsymbol{o}^{(r^\star)},
\boldsymbol{\Omega}_{\boldsymbol{x}},
\boldsymbol{\mathcal{I}}_{\mathrm{use}}^{(r^\star)},
\boldsymbol{f}^{(r^\star)}
\right),
\end{align}
where the components preserve the final opportunity, graph-grounded bottleneck, transferred mechanisms, and creativity--feasibility feedback, respectively.
ROD thereby connects field-level diagnosis to a target-specific and falsifiable opportunity whose motivation, mechanism, and validation path remain traceable to the scholarly structure.

\subsubsection{Innovation assessment}
\label{sec:ia}
Given a research idea or proposal $\boldsymbol{x}$, Innovation Assessment (IA) first decomposes it into assessable claims under a shared analytical schema.
The shared paper and author projections provide relevant evidence and publication-grounded expertise, while review records associated with related papers supply field-specific evaluation norms.
These contexts jointly support idea-specific rubrics, reviewer-conditioned assessments, and actionable revision guidance.

\paragraph{Idea decomposition and evidence grounding}
An LLM agent decomposes $\boldsymbol{x}$ under four analytical aspects:
\begin{align}
\boldsymbol{\mathcal{L}}
=
\{\mathrm{motivation},\mathrm{method},\mathrm{experiment},\mathrm{impact}\}.
\end{align}
For each aspect $\ell\in\boldsymbol{\mathcal{L}}$, it extracts retrieval-oriented claims:
\begin{align}
\boldsymbol{\mathcal{C}}_{\ell}(\boldsymbol{x})
=
\left\{
\boldsymbol{c}_{\ell,i}
\right\}_{i=1}^{n_\ell}.
\end{align}
This shared schema separates the scientific functions of an idea before comparison with prior work.

The shared paper projection produces an evidence pool $\boldsymbol{\mathcal{P}}_{\mathrm{ia}}=\boldsymbol{\mathcal{P}}_{\mathrm{rec}}(\boldsymbol{x})$.
For each paper $\boldsymbol{p}\in\boldsymbol{\mathcal{P}}_{\mathrm{ia}}$, we segment its text into aspect-aware passages $\boldsymbol{\mathcal{D}}_{\ell}(\boldsymbol{p})$ and match each claim only against passages assigned to the same aspect:
\begin{align}
\boldsymbol{\mathcal{N}}
\left(
\boldsymbol{c}_{\ell,i}
\right)
=
\mathrm{TopK}_{\boldsymbol{d}\in
\bigcup_{\boldsymbol{p}\in\boldsymbol{\mathcal{P}}_{\mathrm{ia}}}
\boldsymbol{\mathcal{D}}_{\ell}(\boldsymbol{p})}
\ \mathrm{sim}
\left(
\boldsymbol{c}_{\ell,i},
\boldsymbol{d}
\right).
\end{align}
An alignment agent labels each claim--evidence pair as supported, partially supported, contradicted, or not addressed, while recording their shared points, differences, and missing evidence.
The resulting grounding set is
\begin{align}
\boldsymbol{\mathcal{G}}_{\mathrm{ia}}(\boldsymbol{x})
=
\left\{
\left(
\ell,
\boldsymbol{c}_{\ell,i},
\boldsymbol{d},
\boldsymbol{z}_{\ell,i,d}
\right)
\mid
\ell\in\boldsymbol{\mathcal{L}},
\boldsymbol{c}_{\ell,i}\in\boldsymbol{\mathcal{C}}_{\ell}(\boldsymbol{x}),
\boldsymbol{d}\in\boldsymbol{\mathcal{N}}(\boldsymbol{c}_{\ell,i})
\right\},
\end{align}
where $\boldsymbol{z}_{\ell,i,d}$ contains the alignment label and its evidence-grounded analysis.
This aspect-aware grounding provides a consistent evidential basis for subsequent assessment.

\paragraph{Expert perspective construction}
The shared author projection provides an expert candidate pool $\boldsymbol{\mathcal{A}}_{\mathrm{ia}}=\boldsymbol{\mathcal{A}}_{\mathrm{rec}}(\boldsymbol{x})$ ranked by graph-derived relevance to the idea.
We exclude candidates without sufficient topic-relevant publication evidence and select a reviewer panel $\boldsymbol{\mathcal{R}}_{\mathrm{ia}}$ that balances relevance with diversity in expertise distance.
For each selected author $\boldsymbol{a}$, we retrieve the publications most relevant to $\boldsymbol{x}$ through authorship relations and construct a reviewer profile $\boldsymbol{\pi}_a$ containing major research directions, technical expertise, and representative evidence.
These profiles provide complementary evaluative perspectives grounded in demonstrated publication records; they do not claim to reproduce the selected researchers' personal judgments.

\paragraph{Rubric synthesis from review norms}
Because disciplines differ in their expectations for novelty, soundness, validation, risk, and sufficient evidence, IA constructs an idea-specific rubric rather than applying fixed universal criteria.
From the evidence pool, we identify related papers with accessible review records:
\begin{align}
\boldsymbol{\mathcal{P}}_{\mathrm{rev}}
=
\left\{
\boldsymbol{p}\in\boldsymbol{\mathcal{P}}_{\mathrm{ia}}
\mid
\mathrm{Review}(\boldsymbol{p})\neq\varnothing
\right\}.
\end{align}
For each aspect $\ell\in\boldsymbol{\mathcal{L}}$, an agent extracts historical standards $\boldsymbol{\mathcal{H}}_{\ell}(\boldsymbol{x})$ from the paper content and review comments in $\boldsymbol{\mathcal{P}}_{\mathrm{rev}}$.
Each standard records an evaluation dimension, its underlying concern, and the evidence expected by reviewers.
These field-level norms are combined with $\boldsymbol{\mathcal{G}}_{\mathrm{ia}}(\boldsymbol{x})$, which supplies the boundaries and gaps specific to the input idea:
\begin{align}
\boldsymbol{\mathcal{B}}_{\mathrm{ia}}
=
\operatorname{Rubric}_{\mathrm{agent}}
\left(
\boldsymbol{x},
\boldsymbol{\mathcal{G}}_{\mathrm{ia}}(\boldsymbol{x}),
\left\{
\boldsymbol{\mathcal{H}}_{\ell}(\boldsymbol{x})
\right\}_{\ell\in\boldsymbol{\mathcal{L}}}
\right).
\end{align}
The resulting aspect-wise rubric standardizes how criteria are constructed while adapting their content and evidence requirements to the idea and field.

\paragraph{Reviewer-profile-conditioned assessment}
We organize $\boldsymbol{\mathcal{G}}_{\mathrm{ia}}(\boldsymbol{x})$ into an evidence bank $\boldsymbol{\mathcal{E}}_{\mathrm{ia}}$ containing claim-level evidence cards, paper metadata, citations, and alignment labels.
For each reviewer profile $\boldsymbol{\pi}_a$ and aspect $\ell$, we select the corresponding rubric items $\boldsymbol{\mathcal{B}}_{\ell}$ and evidence cards $\boldsymbol{\mathcal{E}}_{\ell}$ and generate
\begin{align}
\boldsymbol{\rho}_{a,\ell}
=
\operatorname{Assess}_{\mathrm{agent}}
\left(
\boldsymbol{x},
\boldsymbol{\pi}_{a},
\boldsymbol{\mathcal{B}}_{\ell},
\boldsymbol{\mathcal{E}}_{\ell}
\right).
\end{align}
Each $\boldsymbol{\rho}_{a,\ell}$ contains an evidence-linked judgment, strengths, weaknesses, missing evidence or experiments and revision suggestions.
The reviewer-conditioned assessments are consolidated into a meta-assessment $\boldsymbol{\mathcal{M}}_{\mathrm{ia}}(\boldsymbol{x})$ that identifies reviewer consensus and disagreements, shared strengths and weaknesses, and prioritized revision advice:
\begin{align}
\boldsymbol{\mathcal{V}}(\boldsymbol{x})
=
\left(
\left\{
\boldsymbol{\rho}_{a,\ell}
\mid
\boldsymbol{a}\in\boldsymbol{\mathcal{R}}_{\mathrm{ia}},
\ell\in\boldsymbol{\mathcal{L}}
\right\},
\boldsymbol{\mathcal{M}}_{\mathrm{ia}}(\boldsymbol{x})
\right).
\end{align}
Although the evidence, expertise, and evaluation norms vary across ideas and disciplines, IA acquires and organizes them through the same claim-grounding, profile-construction, rubric-synthesis, and assessment process.


\clearpage

\newpage

\normalem
\bibliographystyle{unsrt}
\bibliography{ic}

\clearpage
\section*{Appendix}
\appendix

\section{Details of \textsc{SciAtlasEval} Construction}
\label{app:dataset}
\subsection{Construction of \textsc{SciAtlasEval-HQR}}
\label{app:dataset-hqr}
Heterogeneous Query-Object Retrieval (HQR) evaluates masked retrieval across papers, concepts, and authors.
We use three query object types, namely paper, concept, and author, and two target object types, namely paper and author.
Their combinations define six subtasks: paper-to-paper, paper-to-author, concept-to-paper, concept-to-author, author-to-paper, and author-to-author.
In {\kg}, concepts are represented as keyword nodes.
The relevant relation types include paper-keyword links, authorship links, citation links, and co-authorship links.

For each HQR case, we first select a source node and construct the gold target set from the full {\kg} graph.
Gold targets are constructed from directly connected scholarly objects in the full graph.
For each case, the relation edges that directly connect the query to its gold targets, or otherwise immediately reveal the projected target, are recorded as \texttt{heldout\_edges} and masked during retrieval, while the remaining heterogeneous paths are retained as indirect recovery evidence.
For example, concept-to-paper cases use papers strongly associated with the queried concept.
Author-to-paper cases use controlled holdouts of author-paper relations.
Paper-to-author cases identify relevant authors through papers connected to the query by citation or shared concepts, while masking direct authorship evidence from the query paper.
Each case retains at most 20 gold targets and is required to contain at least five valid gold targets.

This design forces retrieval methods to recover targets through indirect heterogeneous evidence, rather than reading out the most direct relation from the graph.
The final HQR benchmark contains 200 retrieval cases.
The task distribution is 30 paper-to-paper cases, 45 paper-to-author cases, 35 concept-to-paper cases, 30 concept-to-author cases, 40 author-to-paper cases and 20 author-to-author cases.
Each method outputs a ranked list of up to 20 targets.
Performance is evaluated using Recall@20, nDCG@20 and MRR.

\subsection{Construction of \textsc{SciAtlasEval-FCR}}
\label{app:dataset-fcr}
Future Context Recovery (FCR) evaluates whether retrieved historical papers can recover future scientific context.
Each query corresponds to a historical paper and is represented by its title and abstract.
All retrievable papers are restricted to publications available no later than 2022.
Retrieved historical papers are then projected to future papers through citation links from papers published between 2023 and 2025.
The task asks whether the retrieved historical context is sufficient to recover future papers that later connect to the query paper.

We construct FCR from paper-level candidates in {\kg}.
Candidate query papers are required to have a non-empty title and abstract, and to have sufficient future citation support within the 2023--2025 window.
We exclude candidates that are too sparse to support future-context recovery.
We also exclude papers that are too dominant in the citation graph, because such cases can be solved by popularity rather than contextual retrieval.
For each remaining candidate, we build a historical neighbourhood using citation and shared-keyword evidence.
We then compute the future papers reachable from this neighbourhood under a citation-only projection protocol.
Only the subset of future gold papers recoverable under this protocol is used as the final evaluation target.
This makes the benchmark fairer than asking methods to recover future papers that have no recoverable historical evidence under the projection design.

The final FCR benchmark contains 70 paper queries selected from a larger candidate pool according to query quality, recoverable gold validity, projection stability, and difficulty.
For each query, methods return up to 100 historical papers published in or before 2022.
The query paper itself is excluded from projection sources to prevent trivial recovery.
The retrieved historical papers are projected to future papers using the shared citation-only protocol.
The resulting future ranking is evaluated against the recoverable future gold set using Recall@20, nDCG@20, and MRR.
This construction isolates the central ability measured by FCR: retrieving historical papers that are genuinely useful for reconstructing a query paper's future scientific context.

\subsection{Construction of \textsc{SciAtlasEval-Idea}}
\label{app:dataset-idea}
\textsc{SciAtlasEval-Idea} is constructed to evaluate whether scientific AI systems can operate on research ideas as structured scientific objects, rather than only retrieve papers through surface-level textual similarity.
Each instance is a paper-grounded research idea with structured content and associated evidence, allowing us to evaluate idea-level retrieval, contextualization, and innovation assessment.
The dataset contains 349 research ideas, each derived from one source paper.
We construct \textsc{SciAtlasEval-Idea} through three stages.

\paragraph{Paper collection}
We first sample source papers from the disciplinary coverage of {\kg}.
The sampling strategy is designed to avoid concentrating the dataset in a small number of topics, venues, or highly visible research communities.
In addition to broad disciplinary coverage, we explicitly consider venue diversity.
For journals, venue strata are informed by Journal Citation Reports\footnote{\url{https://jcr.clarivate.com/jcr/browse-journals}.}; for conferences, they are informed by ICORE Conference Rankings\footnote{\url{https://portal.core.edu.au/conf-ranks/}.}.
We additionally sample rejected and unpublished works from preprint repositories.
This procedure produces a source-paper pool that includes both high-impact and less prominent venues.
The resulting dataset therefore tests whether a system can assess research ideas across different levels of publication maturity and visibility, rather than only recognizing ideas from a narrow set of well-known papers.

\paragraph{Idea extraction}
For each selected source paper, we use an extraction agent based on \textsc{GPT-5.5} to convert the paper into a structured research idea.
The agent extracts four components: motivation, method, experiment, and impact.
The motivation component describes the research problem and the gap addressed by the paper.
The method component summarizes the main technical, empirical, or conceptual approach.
The experiment component records the principal validation setting and supporting evidence.
The impact component captures the intended contribution or potential significance of the work.
We also identify core reference papers from the source paper's bibliography.
These references are selected when they directly motivate, support, or contextualize the extracted idea.
They provide paper-level evidence for later retrieval and assessment experiments.

\paragraph{Human review}
Human experts then review the extracted ideas for faithfulness and completeness.
The review checks whether each component is supported by the source paper, whether the four components form a coherent research idea, and whether the selected references are appropriate contextual evidence.
Instances with incomplete, ambiguous, or weakly grounded extractions are revised or removed.
This review step is intended to ensure that \textsc{SciAtlasEval-Idea} evaluates idea understanding and assessment, rather than noise introduced by automatic extraction.

\section{Details of Evaluation Metrics}
\label{app:metrics}

This section details the evaluation protocols used across the evaluated datasets and benchmarks.

\subsection{Evaluation metrics for SurveyLens}
\label{app:metrics-surveylens}
SurveyLens evaluates generated survey-style outputs from three complementary perspectives: discipline-specific writing quality, redundancy-aware coverage of a human reference, and local semantic alignment with the reference survey. Following SurveyLens~\cite{surveylens}, each survey is represented as $\mathcal{S}=(\mathcal{O},\mathcal{C},\mathcal{R})$, where $\mathcal{O}$ is the outline, $\mathcal{C}$ is the content, and $\mathcal{R}$ is the reference list.

\paragraph{Discipline-aware rubric score}
This metric measures whether a generated survey follows the writing standards of its target discipline. For each discipline $d$ and survey component $c\in\{\mathcal{O},\mathcal{C},\mathcal{R}\}$, SurveyLens constructs discipline-specific rubrics from human-written surveys. An LLM judge assigns an aspect score $x_{i,k}$ on a 1--5 scale for survey $s_i$ and rubric aspect $k$. The score is normalized as $\bar{x}_{i,k}=x_{i,k}/5$. Aspect weights are learned with a Bradley--Terry model from pairwise preferences and normalized as $\tilde{w}_{d,c,k}$. The component-level rubric score is
\begin{equation}
\mathcal{S}^{\mathrm{rub}}_{d,c}(s_i)
=
5\sum_k \tilde{w}_{d,c,k}\bar{x}_{i,k}.
\end{equation}
Higher values indicate better compliance with discipline-specific expectations for outline structure, content synthesis, and reference use. We report the macro-average over evaluated topics and disciplines.

\paragraph{Redundancy-aware alignment F1}
RA-AlignF1 measures whether a generated survey covers the human-written reference without inflating the output through repetitive content. For each component-topic pair $(c,t)$, the generated survey and reference survey are split into entry sets $\mathcal{E}_{c,t}$ and $\mathcal{G}_{c,t}$. Entries are matched one-to-one using the Hungarian algorithm based on semantic similarity. Given the matched set $\mathcal{M}$, threshold $\tau_c$, and redundancy weight $\omega(e)$ for generated entry $e$, precision and recall are
\begin{equation}
P=
\frac{1}{|\mathcal{E}_{c,t}|}
\sum_{(e,g)\in\mathcal{M}}
\omega(e)\mathbb{I}[\mathrm{sim}(e,g)\geq\tau_c],
\quad
R=
\frac{1}{|\mathcal{G}_{c,t}|}
\sum_{(e,g)\in\mathcal{M}}
\mathbb{I}[\mathrm{sim}(e,g)\geq\tau_c].
\end{equation}
The redundancy weight is
\begin{equation}
\omega(e)=
\exp\left(
-\lambda
\max_{e'\in\mathcal{E}_{c,t}\setminus\{e\}}
\mathrm{sim}(e,e')
\right),
\end{equation}
which down-weights generated entries that are highly similar to other generated entries. The final score is
\begin{equation}
\mathrm{RA\text{-}AlignF1}=\frac{2PR}{P+R}.
\end{equation}
Higher RA-AlignF1 indicates stronger reference coverage with less redundant generation.

\paragraph{Thresholded MaxSim}
$\tau$-MaxSim measures local semantic alignment between generated entries and human-reference entries without enforcing one-to-one matching. For component $c$, let $\mathcal{E}_{c}=\bigcup_t\mathcal{E}_{c,t}$ and $\mathcal{G}_{c}=\bigcup_t\mathcal{G}_{c,t}$. The score is
\begin{equation}
\tau\text{-}\mathrm{MaxSim}
=
\frac{1}{|\mathcal{E}_{c}|}
\sum_{e\in\mathcal{E}_{c}}
\max\left\{
0,
\max_{g\in\mathcal{G}_{c}}\mathrm{sim}(e,g)-\tau_c
\right\}.
\end{equation}
Unlike RA-AlignF1, this metric does not penalize repeated matches to the same reference entry. It captures whether generated entries contain locally relevant semantic content, whereas RA-AlignF1 better reflects global coverage and redundancy control.

\subsection{Evaluation metrics for LiveIdeaBench}
\label{app:metrics-liveideabench}
LiveIdeaBench evaluates scientific idea generation from single-keyword prompts.
Following the original protocol \cite{liveideabench}, we use a fixed ten-model judge panel selected from LiveBench \cite{livebench} under organizational-diversity and base-model deduplication constraints: Claude 3.7 Sonnet Thinking, o3-mini-high, GPT-4.5-preview, QwQ-32B, DeepSeek-R1, Gemini 2.0 Flash Thinking, Gemini 2.0 Pro, Qwen-Max, DeepSeek-V3, and Claude 3.5 Sonnet.
When a generating model belongs to the panel, it is excluded from the eligible judges to prevent self-evaluation.
For each generated idea, originality, feasibility and clarity are scored by three judges randomly sampled from the remaining eligible panel and averaged across judges.
Fluency is evaluated by one randomly sampled eligible judge for each keyword--system pair.
1) \textit{Originality} measures whether an idea is novel rather than a direct restatement or minor variation of existing work. Each idea is scored by three randomly sampled eligible judges, and the final score is averaged over judges, generated ideas, and keywords.
2) \textit{Feasibility} measures whether the idea is technically plausible, scientifically sound, and practically implementable. It uses the same three-judge scoring protocol as originality, with scores averaged over all generated ideas and keywords.
3) \textit{Clarity} measures whether the idea is expressed as a coherent and understandable research proposal. Judges assess whether the motivation, proposed direction, and expected contribution can be clearly interpreted from the generated text.
4) \textit{Fluency} measures the diversity of ideas generated for the same keyword. A judge compares the set of ideas produced by a method for a given keyword and assigns one of four grades: D for academically identical ideas, C for similar ideas addressing similar problems, B for different ideas addressing similar problems, and A for completely different ideas addressing different problems. These grades are mapped to scores of 1, 4, 7, and 10, respectively, and averaged over keywords.
5) \textit{Flexibility} measures whether a method maintains reliable performance across different keywords and scientific domains. For each keyword, LiveIdeaBench first averages originality, feasibility, clarity, and fluency. The flexibility score is then defined as the 30th percentile of these keyword-level composite scores, reflecting the method's performance floor rather than only its best cases.

\subsection{Evaluation metrics for \textsc{SciAtlasEval-Idea}}
\label{app:metrics-idea}
\textsc{SciAtlasEval-Idea} supports two types of evaluation: idea-assessment report evaluation and context-retrieval evaluation. The former measures whether an assessment report provides reliable and actionable judgments for a research idea. The latter measures whether a method can retrieve useful papers and expert authors for grounding the assessment.

\subsubsection{Idea-assessment report evaluation}
\label{app:metrics-idea-ass}
For each idea, we compare the assessment report generated by {\kg} with the report generated by a baseline method. Following the qualitative evaluation protocol in InnoEval \cite{innoeval}, LLM judges perform pairwise comparison along five dimensions: \textit{Rationality}, which measures whether the report is logically sound; \textit{Supportiveness}, which measures whether its claims are backed by concrete evidence; \textit{Depth}, which measures whether the analysis goes beyond surface-level comments; \textit{Constructiveness}, which measures whether the report provides useful and feasible revision suggestions; and \textit{Overall Quality}, which gives a holistic judgment of the complete report.
For each dimension, the judge selects the better report between {\kg} and a baseline, with ties permitted when neither report has a clear advantage.
A win for {\kg} receives a score of $1$, whereas a tie or loss receives a score of $0$.
For each dimension and baseline, the win rate is
\begin{equation}
\mathrm{Win\ Rate}
=
\frac{N_{\mathrm{win}}}
{N_{\mathrm{win}}+N_{\mathrm{tie}}+N_{\mathrm{loss}}},
\end{equation}
where the counts aggregate all idea--judge comparisons for the corresponding dimension and baseline.

\subsubsection{Paper-retrieval evaluation}
\label{app:metrics-idea-paper}
For each idea $I$, a method retrieves a paper set $R=\{r_i\}_{i=1}^{|R|}$.
Following InnoEval \cite{innoeval}, let $R_h\subseteq R$ denote the subset of retrieved papers judged by an LLM as highly relevant. Let $T_I$ denote the main topic words in idea $I$, and $T_{r_i}$ denote the main topic words in retrieved paper $r_i$.

\paragraph{Relevance density}
Relevance density measures how concentrated the retrieved set is with highly relevant papers:
\begin{equation}
\mathrm{Relevance\ Density}
=
\frac{|R_h|}{|R|}.
\end{equation}
This prevents a method from receiving a high score merely by retrieving many papers.

\paragraph{Topic coverage}
Topic coverage measures how much of the idea's topic space is covered by the retrieved papers:
\begin{equation}
\mathrm{Topic\ Coverage}
=
\frac{
\left|
T_I
\cap
\left(
T_{r_1}\cup\cdots\cup T_{r_{|R|}}
\right)
\right|
}{
|T_I|
}.
\end{equation}
Higher values indicate that the retrieved papers cover more key concepts in the input idea.

\paragraph{Diversity}
Diversity measures how much additional topic information is introduced by the retrieved papers beyond the original idea:
\begin{equation}
\mathrm{Diversity}
=
\frac{
\left|
\left(
T_{r_1}\cup\cdots\cup T_{r_{|R|}}
\right)
-
T_I
\right|
}{
\left|
T_{r_1}\cup\cdots\cup T_{r_{|R|}}
\right|
}.
\end{equation}
Higher values indicate that the retrieval results provide broader contextual expansion rather than only repeating the idea's existing topic terms.

\paragraph{Quality}
Quality measures the overall usefulness of the retrieved papers. An LLM assigns each retrieved paper $r_i$ a quality score $s_{r_i}$, and the final score is
\begin{equation}
\mathrm{Quality}
=
\frac{1}{|R|}
\sum_{i=1}^{|R|}
s_{r_i}.
\end{equation}

\subsubsection{Author-retrieval evaluation}
\label{app:metrics-idea-author}
For each idea or paper, a method retrieves an author panel $A=\{a_i\}_{i=1}^{m}$, where $m=10$ in our experiments. We evaluate the panel using relevance, diversity, and authority.

\paragraph{Relevance} Relevance measures whether the panel contains at least one author whose expertise is strongly matched to the target idea or paper. Each author is scored by an LLM on a 0--10 scale using the target metadata and the author's academic background. The panel-level relevance score is the best individual match:
\begin{equation}
\mathrm{Relevance}(A)
=
\max_{a_i\in A}
s^{\mathrm{rel}}(a_i).
\end{equation}

\paragraph{Diversity} Diversity measures the breadth of expertise within the author panel. The LLM judge receives only the authors' academic backgrounds, without the target idea or paper, and assigns a 0--10 score based on differences in discipline, methodology, application area, and research perspective. Thus, diversity evaluates complementarity within the panel rather than relevance to the target.

\paragraph{Authority} Authority measures the scholarly strength of the retrieved authors without using an LLM judge. For each author $a_i$, let $p_i$ be the publication count and $c_i$ be the citation count from the author's relevant papers. Let $p_{95}$ and $c_{95}$ be the global 95th percentiles of publication counts and citation counts across all evaluated authors. We compute
\begin{equation}
u_i^{p}
=
\min\left(
1,
\frac{\log(1+p_i)}{\log(1+p_{95})}
\right),
\quad
u_i^{c}
=
\min\left(
1,
\frac{\log(1+c_i)}{\log(1+c_{95})}
\right).
\end{equation}
The individual authority score is
\begin{equation}
s^{\mathrm{auth}}(a_i)
=
10\left(
0.4u_i^{p}
+
0.6u_i^{c}
\right),
\end{equation}
and the panel-level authority score is the mean over all retrieved authors:
\begin{equation}
\mathrm{Authority}(A)
=
\frac{1}{m}
\sum_{i=1}^{m}
s^{\mathrm{auth}}(a_i).
\end{equation}

\subsection{Evaluation metrics for the Chain-of-Ideas refinement dataset}
\label{app:metrics-idea-refinement}
We use the Chain-of-Ideas dataset to evaluate whether an assessment report can improve a research idea when used as feedback. The original Chain-of-Ideas benchmark collects 50 recent AI research topics from Hugging Face Daily Papers and evaluates generated ideas with the Idea Arena protocol~\cite{chain-of-ideas}. In our setting, we take the unoptimized Chain-of-Ideas output as the original idea. Each assessment method generates feedback for the same idea, and the idea is then revised according to that feedback. The evaluation compares the revised ideas against one another and against the original idea.

Following Idea Arena, evaluation is performed by pairwise comparison rather than independent Likert scoring. For each topic, the judge is given two anonymized ideas and selects the better one for each of five dimensions: \textit{Novelty}, which measures whether the problem or approach is new; \textit{Significance}, which measures whether the idea addresses an important problem and is likely to influence future work; \textit{Feasibility}, which measures whether the idea can be realized with existing methods and has no obvious technical flaw; \textit{Clarity}, which measures whether the idea is well organized and understandable; and \textit{Effectiveness}, which measures whether the proposed idea is likely to work better than existing baselines. The judge may also mark a tie. Each pair is evaluated twice with reversed order to reduce position bias.

The resulting win--loss--tie records are converted into ELO scores using the official Idea Arena implementation and its unmodified default settings from Chain-of-Ideas \cite{chain-of-ideas}.
ELO scores are computed independently for novelty, significance, feasibility, clarity, and effectiveness, and their mean is reported as the overall feedback-refinement score.
Higher ELO indicates that ideas revised using a method's feedback are more often preferred in pairwise comparison.

\subsection{Evaluation metrics for \textsc{SciAtlasEval-HQR} and \textsc{FCR}}
\label{app:metrics-hqr-fcr}
\textsc{SciAtlasEval-HQR} and \textsc{FCR} evaluate whether a method can retrieve target scholarly objects from the heterogeneous topology of {\kg}. In Heterogeneous Query-Object Retrieval (HQR), the targets are papers or authors after direct evidence edges are masked.
In Future Context Recovery (FCR), retrieved historical papers are projected to future papers through citation links. Both settings are evaluated with standard information-retrieval metrics.
\textit{Recall@20} measures the fraction of gold targets that appear in the top-20 retrieved results. \textit{nDCG@20} measures ranking quality by giving higher credit when relevant targets appear closer to the top of the ranked list. \textit{MRR} measures how early the first relevant target appears, using the reciprocal rank of the first correct result.
For HQR, we average each metric within each subtask and report the macro-average across the six subtasks; for FCR, we average each metric over all queries.

\subsection{Evaluation metrics for TaxoBench}
\label{app:metrics-taxobench}
TaxoBench evaluates whether a retrieval method can recover the papers associated with an expert-authored survey taxonomy~\cite{taxobench}. For each topic, let $G$ be the gold paper set and $R$ be the retrieved paper set. We report standard precision, recall, and F1.
\textit{Precision} measures the fraction of retrieved papers that are included in the gold set.
\textit{Recall} measures the fraction of gold papers recovered by the retrieved set.
\textit{F1} is the harmonic mean of precision and recall.

\subsection{Evaluation metrics for ScholarQABench}
\label{app:metrics-scholarqabench}
ScholarQABench evaluates scientific question answering and literature synthesis across single-paper and multi-paper settings~\cite{openscholar}. We report the metrics used in the original benchmark for the covered subsets: PubMedQA (Pub), SciFact (Sci), QASA, ScholarQA-CS (CS), ScholarQA-Multi (Multi), ScholarQA-Bio (Bio), and ScholarQA-Neuro (Neu).

\paragraph{Correctness} \textit{Corr} measures whether the generated answer matches the gold answer or expert reference. For PubMedQA and SciFact, correctness is standard accuracy over the predicted answer label. For QASA, correctness is ROUGE-L against the reference answer. For ScholarQA-CS, correctness is computed with expert-written rubrics. Each answer is scored on general criteria, such as length, expertise, citations, and excerpts, and on annotation-driven criteria, which check whether expert-identified key ingredients are present. The final CS correctness score is a weighted sum, with 40\% assigned to general criteria and 60\% to annotation-driven criteria.

\paragraph{Citation accuracy} \textit{Cite} measures whether generated citations correctly support citation-worthy statements. Following ScholarQABench, citation quality is evaluated at the sentence level with an attribution model. Citation recall checks whether each citation-worthy statement has appropriate supporting citations. Citation precision checks whether each provided citation supports the associated statement and is necessary. The reported citation score is citation F1:
\begin{equation}
\mathrm{Cite\text{-}F1}
=
\frac{
2\cdot \mathrm{Cite\text{-}P}\cdot \mathrm{Cite\text{-}R}
}{
\mathrm{Cite\text{-}P}+\mathrm{Cite\text{-}R}
}.
\end{equation}

\paragraph{LLM-judged answer quality} \textit{LLM} is used for the multi-paper long-form setting where answer quality cannot be fully captured by exact labels or reference overlap. Following ScholarQABench, an evaluator model scores three aspects: organization, relevance, and coverage. Organization measures whether the answer is logically structured; relevance measures whether it stays focused on the question; and coverage measures whether it contains sufficient information. We report their average:
\begin{equation}
\mathrm{LLM}
=
\frac{
\mathrm{Org}+\mathrm{Rel}+\mathrm{Cov}
}{3}.
\end{equation}
In our experiments, this metric is reported for the ScholarQA-Multi subset, together with citation accuracy.

\section{Baselines}

\subsection{Baselines for foundational scientific retrieval}
\label{app:baselines-foundational}
For the foundational retrieval and question-answering experiments in Fig.~\ref{fig:main_results}, we compare {\kg} with a set of lexical, embedding-based, scientific-document, retrieval-augmented, and agentic systems.
\begin{itemize}
    \item \textbf{BM25}. BM25 is a classical sparse lexical retrieval method that ranks documents by exact term overlap with inverse-document-frequency and document-length normalization. We use it as the lexical retrieval baseline. Except for ScholarQABench, the searchable corpus is built from the paper nodes in {\kg}, and each candidate paper is represented by its title and abstract. For ScholarQABench, we follow the OpenScholar setting, where BM25 retrieves evidence documents from the OpenScholar datastore.

    For paper-like queries, including paper queries in HQR and FCR, idea descriptions in \textsc{SciAtlasEval-Idea}, and topic queries in TaxoBench, we use the available title, abstract, topic statement, or structured idea text as the lexical query. For concept queries, the concept phrase is used directly. For author queries, the query is expanded from the author name to a lightweight profile consisting of the author name and representative observed paper titles, because BM25 has no author representation, and a name alone provides little topical signal. For scientific question-answering, the natural-language question is used to retrieve evidence passages from the OpenScholar datastore.

    The retrieved objects are then adapted according to the target type. For paper targets, BM25 directly ranks candidate papers by lexical matching. For author targets, BM25 does not retrieve authors natively; its paper results are therefore converted to authors through the shared paper-to-author projection used for all text-to-paper baselines. For question-answering targets, BM25 serves only as the evidence retriever, and the retrieved evidence is passed to the same answer-generation and evaluation pipeline as defined in OpenScholar.

    \item \textbf{BGE}~\cite{bge}. BGE is a dense embedding retrieval baseline from the BAAI General Embedding family. It encodes queries and candidate texts into vector representations and ranks candidates by semantic similarity. Except for ScholarQABench, the searchable corpus is built from the paper nodes in {\kg}, and each candidate paper is represented by its title and abstract embeddings. For ScholarQABench, we follow the OpenScholar setting, where BGE retrieves evidence documents from the OpenScholar datastore.

    For paper-like queries, including paper queries in HQR and FCR, idea descriptions in \textsc{SciAtlasEval-Idea}, and topic queries in TaxoBench, we encode the available title, abstract, topic statement, or structured idea text as the retrieval query. For concept queries, the concept phrase is encoded directly. For author queries, the query is expanded from the author name to a lightweight profile consisting of the author name and representative observed paper titles, so that the embedding retriever receives semantic information about the author's research area rather than only a name string. For scientific question-answering, the natural-language question is encoded and used to retrieve evidence passages or documents from the OpenScholar datastore.

    The retrieved objects are then adapted according to the target type. For paper targets, BGE ranks candidate papers by query--paper embedding similarity. For author targets, BGE does not retrieve authors natively; its paper results are therefore converted to authors through the shared paper-to-author projection used for all text-to-paper baselines. For question-answering targets, BGE serves only as the dense evidence retriever, and the retrieved evidence is passed to the same answer-generation and evaluation pipeline used in OpenScholar.

    \item \textbf{SPECTER2}~\cite{specter2}. SPECTER2 is a scientific document embedding baseline designed specifically for scholarly papers. Unlike BGE, which is a general-purpose text embedding model, SPECTER2 represents papers in a scientific-document embedding space learned from paper text and citation-related supervision, and uses a query adapter for retrieval. We use it to test whether a paper-specialized semantic representation can recover the relevant scientific context without using the heterogeneous graph structure of {\kg}.

    For paper-like, concept, and author-profile queries, the query text is encoded with the SPECTER2 query encoder and matched against SPECTER2 embeddings of candidate papers. In our implementation, SPECTER2 retrieves from the local SciAtlas paper corpus through the SPECTER2 paper embedding index; in FCR, the same historical cutoff is applied before evaluation. Because SPECTER2 returns papers rather than authors or answers, its outputs are adapted in the same way as other text-to-paper baselines: paper targets are evaluated directly, author targets use the shared paper-to-author projection, and ScholarQABench uses SPECTER2 only as the evidence retriever before the OpenScholar answer-generation and evaluation pipeline.

    \item \textbf{OpenScholar}~\cite{openscholar}. OpenScholar is a retrieval-augmented system designed for scientific literature synthesis. Its original pipeline combines a large open-access scientific datastore, a trained retriever and reranker, and an answer generator for citation-backed synthesis. In our non-ScholarQABench retrieval experiments, we use only the OpenScholar retrieval component as a literature-specialized retriever, not its full answer-generation pipeline. This tests whether a retriever trained for scientific literature synthesis can recover task-relevant papers without using the heterogeneous graph structure of {\kg}.

    For paper-like, concept, and author-profile queries, the query text is submitted to the OpenScholar retrieval runtime, which retrieves candidates from its own literature datastore rather than directly from the SciAtlas paper corpus. In FCR, the same historical cutoff is applied after mapping. Because OpenScholar returns papers or evidence documents rather than authors, author targets use the shared paper-to-author projection. For ScholarQABench, we use OpenScholar in its native setting, including its datastore, retriever, and answer-generation pipeline under the original evaluation protocol.

    \item \textbf{PaperQA2}~\cite{paperqa2}. PaperQA2 is an agentic retrieval-augmented system for scientific literature search, evidence gathering and cited answer generation. It searches for papers, gathers relevant evidence chunks, reranks or summarizes evidence, and generates grounded answers with citations. We include PaperQA2 only in the ScholarQABench question-answering evaluation, where the task requires answer generation rather than only ranked paper retrieval.

    \item \textbf{Claude Code}. Claude Code is an agentic system that can read task context, plan actions, and use tools such as command-line or search workflows. We use it as an agentic deep search baseline. Given a scientific query, paper description, or author query, Claude Code can generate scholarly search expressions and return candidate objects through external search tools. It can also self-refine the search query based on the searched results. For each dataset, we clearly define the task object in its prompt and enable it to iterate automatically by using its own agentic capability. For ScholarQABench, we enable it to self-organize the answer based on the searched results.

    \item \textbf{Naive author aggregation}. This baseline tests a simple paper-first strategy for author retrieval. It first retrieves topically similar papers and then ranks authors by aggregating the authorship evidence from those papers. It does not model author communities, expertise diversity or authority beyond the retrieved paper set. In our author-retrieval evaluation, it receives the same target idea or paper metadata as {\kg}, retrieves candidate papers, projects them to authors, and is scored with the same relevance, diversity, and authority metrics.

    \item \textbf{OpenReview expertise}. The OpenReview expertise baseline follows the reviewer-paper affinity setting used in OpenReview reviewer matching, where reviewer expertise is estimated from publication profiles using retrieval or representation models such as BM25, SPECTER, and multifacet recommendation. We adapt it as a paper-author retrieval baseline by treating the paper as the submission and candidate authors as reviewer-like experts.

    \item \textbf{{\kg} w/o Keyword}. This ablation removes the keyword layer and paper--keyword evidence from {\kg}. It tests whether fine-grained conceptual anchors are necessary for retrieving scientifically related papers that may not share surface terms.

    \item \textbf{{\kg} w/o Topological Propagation}. This ablation disables topological propagation over the graph. Retrieval therefore relies on direct anchoring signals without multi-hop diffusion through citation, concept, and expertise relations. It tests whether the gains of {\kg} come from graph propagation rather than only from better initial matching.
\end{itemize}

\subsection{Baselines for Research Development Trajectory}
\label{app:baselines-rdt}
All baselines in this experiment follow the implementation and evaluation setting provided by SurveyLens~\cite{surveylens}. Each system receives the same survey topic input and is evaluated under the same SurveyLens discipline-aware rubric and canonical-alignment metrics.

\begin{itemize}
    \item \textbf{AutoSurvey}~\cite{autosurvey}. AutoSurvey is an automated survey-generation pipeline built around retrieval-augmented outline construction and section-wise drafting. It first retrieves relevant papers for a given topic and generates a structured outline, then drafts subsections in parallel with LLMs, merges and refines the generated sections, and selects or improves the final survey through an evaluation-and-iteration stage.

    \item \textbf{SurveyForge}~\cite{surveyforge}. SurveyForge focuses on improving outline quality and citation reliability in automatic survey writing. It generates outlines by drawing on heuristics from human-written survey structures and retrieved domain papers, then uses a memory-driven scholar navigation agent to retrieve high-quality references for content generation and refinement.

    \item \textbf{AutoSurvey2}~\cite{autosurvey2}. AutoSurvey2 extends the automated survey-generation pipeline with retrieval-augmented synthesis, parallel section generation, and iterative refinement. It emphasizes real-time retrieval of recent literature, structured long-form generation, and automated multi-LLM evaluation to improve topical completeness, structural coherence, and citation fidelity.

    \item \textbf{InteractiveSurvey}~\cite{interactivesurvey}. InteractiveSurvey is an LLM-based personalized survey-generation system. It supports survey construction from retrieved or user-provided references and organizes the process around reference categorization, outline generation, and content drafting. Its original design allows users to interactively refine intermediate outputs, including categories, outlines, and generated sections.

    \item \textbf{SurveyX}~\cite{surveyx}. SurveyX models survey writing as a two-stage process consisting of preparation and generation. It uses online reference retrieval, an AttributeTree-style preprocessing structure to organize the literature, and a re-polishing step to improve the final survey. The method is designed to produce more organized surveys with stronger content and citation quality.

    \item \textbf{SciSage}~\cite{scisage}. SciSage is a multi-agent survey-generation framework built around a ``reflect-when-you-write'' strategy. It combines agents for query interpretation, content retrieval, drafting, and refinement with a hierarchical reflector that reviews drafts at the outline, section, and document levels, aiming to improve structural coherence, depth of analysis, and citation quality.
\end{itemize}

\subsection{Baselines for Research Opportunity Discovery}
\label{app:baselines-opportunity}
All baselines in this experiment are reproduced following their original papers or public implementations whenever available. The generated research ideas are evaluated under the LiveIdeaBench setting~\cite{liveideabench}, using the same keyword inputs and the same evaluation dimensions.

\begin{itemize}
    \item \textbf{Naive LLM}. This baseline directly prompts the backbone LLM to generate research ideas from the input keyword. It provides a minimal prompting baseline for testing how much opportunity discovery can be achieved from the model's parametric knowledge alone.

    \item \textbf{Chain-of-Ideas}~\cite{chain-of-ideas}. Chain-of-Ideas organizes related literature into an evolving chain that represents the progressive development of a research area. The LLM then extrapolates from this chain to propose a new research idea and experimental direction. It is included as a literature-evolution baseline that explicitly models how prior ideas develop over time before generating a candidate opportunity.

    \item \textbf{Claude Code}. Claude Code is used as an agentic idea-generation baseline. Given the keyword-level task input, it can plan intermediate search or reasoning steps and synthesize a compact research proposal. Compared with the naive LLM baseline, this setting tests whether a general-purpose agentic workflow can improve opportunity discovery without using the structured scholarly topology of {\kg}.

    \item \textbf{Co-Scientist}~\cite{co-scientist}. Co-Scientist is a multi-agent framework that generates, debates, ranks, and evolves scientific hypotheses. Because the original AI-CoScientist system does not provide open-source code, we use the open-source reproduction \texttt{jataware/open-coscientist}\footnote{\url{https://github.com/jataware/open-coscientist}} for implementation. This baseline represents a hypothesis-generation system based on multi-agent proposal and self-improvement.

    \item \textbf{ResearchAgent}~\cite{researchagent}. ResearchAgent generates research ideas starting from a core reference paper, augments it with related publications and concept-level knowledge, and iteratively refines the proposed problem, method, and experiment through LLM-based reviewing agents. Since ResearchAgent requires reference papers as input, we follow the original setting by constructing a reference-paper set for each LiveIdeaBench topic before idea generation.

    \item \textbf{AI-Scientist}~\cite{ai-scientist}. AI-Scientist is an automated scientific-discovery framework that generates ideas, designs experiments, executes code, analyzes results, and writes papers in its full setting. In this experiment, we use its idea-generation stage as a baseline for opportunity discovery. It represents an AI-scientist-style agent that proposes research directions through self-reflection and novelty-oriented reasoning.
\end{itemize}

\subsection{Baselines for Innovation Assessment}
\label{app:baselines-inno-ass}
The innovation-assessment experiments include two settings: report-quality evaluation and feedback-guided idea refinement. CoT and RAG are implemented by us as simple prompting and retrieval-augmented prompting baselines. Other specialized baselines follow the settings of their original papers. In the idea-refinement experiment, the assessment report generated by each baseline is used as feedback in the novelty-check stage of Chain-of-Ideas, and the revised ideas are then evaluated under the same pairwise ELO protocol.

\begin{itemize}
    \item \textbf{CoT}. CoT is a direct reasoning baseline. Given the structured research idea, the model is prompted to generate an innovation-assessment report by reasoning over the idea itself, without external literature retrieval or expert-profile construction. This baseline tests whether the backbone model can judge novelty, soundness, evidence support, and revision potential from the idea text alone.

    \item \textbf{RAG}. RAG extends the CoT baseline with retrieved literature evidence. For each input idea, we retrieve related papers from the SciAtlas paper corpus and provide the retrieved paper context to the model before it generates the assessment report. This baseline tests whether adding local paper-level evidence improves idea assessment, without using the full {\kg} workflow for reviewer selection, norm construction, or graph-structured evidence organization.

    \item \textbf{DeepReviewer}~\cite{deepreview}. DeepReviewer is a paper-review model designed to emulate a structured expert-review process with evidence-based reasoning. We use the released DeepReviewer-14B model as the baseline implementation. Because DeepReviewer is designed to evaluate full papers rather than standalone idea cards, we provide it with the source paper behind each \textsc{SciAtlasEval-Idea} instance, i.e., the paper from which the corresponding idea was constructed. The generated review-style output is then used as the assessment report for the associated idea.

    \item \textbf{NoveltyAgent}~\cite{noveltyagent}. NoveltyAgent is an autonomous novelty-reporting system that decomposes a manuscript into fine-grained novelty points, retrieves related papers for each point, and performs point-wise comparison with self-validation. We use it as a specialized novelty-assessment baseline, focusing on whether the proposed idea is original relative to retrieved prior work.

    \item \textbf{ScholarEval}~\cite{scholareval}. ScholarEval is a literature-grounded research-idea evaluation framework. It assesses an idea along dimensions such as empirical soundness and contribution relative to prior work, using retrieved literature as evidence. We use it as a specialized idea-evaluation baseline that explicitly connects idea assessment to literature support and contribution analysis.
\end{itemize}

\renewcommand\thefigure{\thesection} 
\renewcommand\thetable{\thesection}

\end{document}